\documentclass[11pt]{article}

\usepackage[final]{acl}

\usepackage{times}
\usepackage{latexsym}

\usepackage[T1]{fontenc}

\usepackage[utf8]{inputenc}

\usepackage{microtype}

\usepackage{inconsolata}

\usepackage{graphicx}

\title{Instructions for *ACL Proceedings}

\author{First Author \\
  Affiliation / Address line 1 \\
  Affiliation / Address line 2 \\
  Affiliation / Address line 3 \\
  \texttt{email@domain} \\\And
  Second Author \\
  Affiliation / Address line 1 \\
  Affiliation / Address line 2 \\
  Affiliation / Address line 3 \\
  \texttt{email@domain} \\}

\usepackage[english]{babel}
\usepackage{amsmath}
\usepackage{amssymb}
\usepackage{amsfonts}
\usepackage{graphicx}
\usepackage{epsfig,epsf,rotating,latexsym,amsmath,amssymb,amsfonts,bm,subfigure,epstopdf}
\usepackage{algorithm}
\usepackage{amsthm}
\usepackage[most]{tcolorbox}
\usepackage{algorithm}
\usepackage{algorithmic}
\usepackage{booktabs}
\usepackage{multirow}

\usepackage{newfloat}
\usepackage{listings}
\DeclareCaptionStyle{ruled}{labelfont=normalfont,labelsep=colon,strut=off} 
\floatstyle{ruled}
\newfloat{listing}{tb}{lst}{}
\floatname{listing}{Listing}
\theoremstyle{plain}

\theoremstyle{definition}

\title{GRASP: Generating, Revising, and Assessing for Strategic Planning \\
with Agentic AI}

\author{
\textbf{Arunabh Srivastava\textsuperscript{1}},
\textbf{Mohammad A. (Amir) Khojastepour\textsuperscript{2}},
\textbf{Srimat Chakradhar\textsuperscript{2}},\\
\textbf{Sennur Ulukus\textsuperscript{1}}
\\
\textsuperscript{1}University of Maryland, College Park, MD \\
\textsuperscript{2}NEC Laboratories America, Inc.\\
}

\begin{document}
\maketitle

\begin{abstract}
Large Language Models (LLMs) typically exhibit a performance profile where reliability degrades as task complexity increases. We address the challenge of generating high-quality natural language executable plans for complex tasks by introducing \textbf{GRASP}, a strategy-aware, multi-stage planning framework. GRASP decouples the planning pipeline across specialized, context-isolated modules: it pre-compiles global macro-guidelines (\emph{GenPlan}), explores alternative localized strategies within isolated context windows (\emph{RevPlan}), and independently evaluates trajectories using a multi-criteria discriminator (\emph{VerPlan}). Empirical evaluations show that GRASP consistently establishes a new state-of-the-art frontier across diverse datasets, yielding substantial accuracy gains over direct LLM planners on Natural Plan Calendar Scheduling ($\sim$12.4\%$\uparrow$), ZebraLogic ($\sim$30.8\%$\uparrow$), and SciBench Math. Crucially, under multi-task scaling—where standard planners suffer immediate performance collapse—GRASP completely flattens the multi-task degradation penalty. In interleaved dual-task environments, GRASP achieves an absolute accuracy gain of up to 16.7\% over direct LLM planners. Furthermore, by isolating context and enforcing strict macro-regularization, GRASP outperforms frontier reasoning models (such as GPT-5-mini) by a margin of 14.5\%.
\end{abstract}

\section{Introduction}
Large Language Model (LLM) agents have emerged as a prominent paradigm for automating complex, multi-step workflows across diverse enterprise domains~\cite{wang2025surveyhas}. Unlike traditional deterministic software, LLM-based agents can dynamically interpret open-vocabulary natural language specifications (\emph{tasks}) and adapt to real-time context shifts (\emph{task instances}) without explicit reprogramming~\cite{wooldridge1995intelligent}. However, directly steering an LLM using layered, long-horizon natural language instructions presents severe architectural challenges. As task complexity (as defined in \cite{dziri23faith}) scales, models experience a sharp performance degradation, often termed the ``Curse of Instructions''~\cite{haradacurse}, where constraint collisions and attention fatigue trigger hallucinations and multi-step execution failures~\cite{wei2022chain, yao2023tree}. This vulnerability highlights the need for explicit plan generation that transforms high-level descriptions into structured, decomposed execution paths.

Effective plan generation mitigates this scaling penalty by dividing an intricate task into ordered sub-steps whose individual complexity falls within the model's reliable operating range. In this paper, we propose \textbf{GRASP}, an autonomous framework for asynchronous plan generation in agentic AI systems. Inspired by human cognitive processes—namely identification, reasoning, and judgment—GRASP decouples the planning pipeline into specialized, context-isolated modules. The architecture first compiles global hard constraints and soft guidelines (\emph{GenPlan}), explores localized strategic paths across separate context windows (\emph{RevPlan}), and independently evaluates trajectories using a multi-criteria discriminator (\emph{VerPlan}). This strict contextual separation prevents cross-contamination and breaks the compounding error cascades that degrade single-track planners.

Finally, robust execution is paramount to translating high-quality plans into correct downstream results. Attempting to execute a multi-step plan within a single monolithic LLM track collapses the structural benefits of decomposition and exacerbates context pollution. To address this, we interface GRASP with RunAgent \cite{srivastava2026runagentinterpretingnaturallanguageplans}, a multi-agent plan execution platform.

In summary, the primary contributions of this work are three-fold: (1) We introduce GRASP, an autonomous, multi-stage planning framework that enforces global macro-regularization and context-isolated strategic refinement to eliminate compounding error feedback loops in LLM planning. (2) We demonstrate the critical interdependence of modular planning components through a comprehensive ablation study, revealing how local search optimization collapses into unconstrained hallucination loops without global macro guardrails. (3) Through extensive empirical evaluations spanning programmatic execution and pure LLM execution, we demonstrate that GRASP establishes a new state-of-the-art frontier. Strikingly, under multi-task scaling, GRASP completely neutralizes the contextual degradation penalties that plague standard language architectures. In interleaved dual-task environments, GRASP enabled by standard language backbones achieves an absolute accuracy gain of up to $16.7\%$ over direct baselines, systematically outpacing premium, native frontier reasoning models such as GPT-5-mini by $14.5\%$.

\begin{figure}
    \centering
    \includegraphics[width=\linewidth]{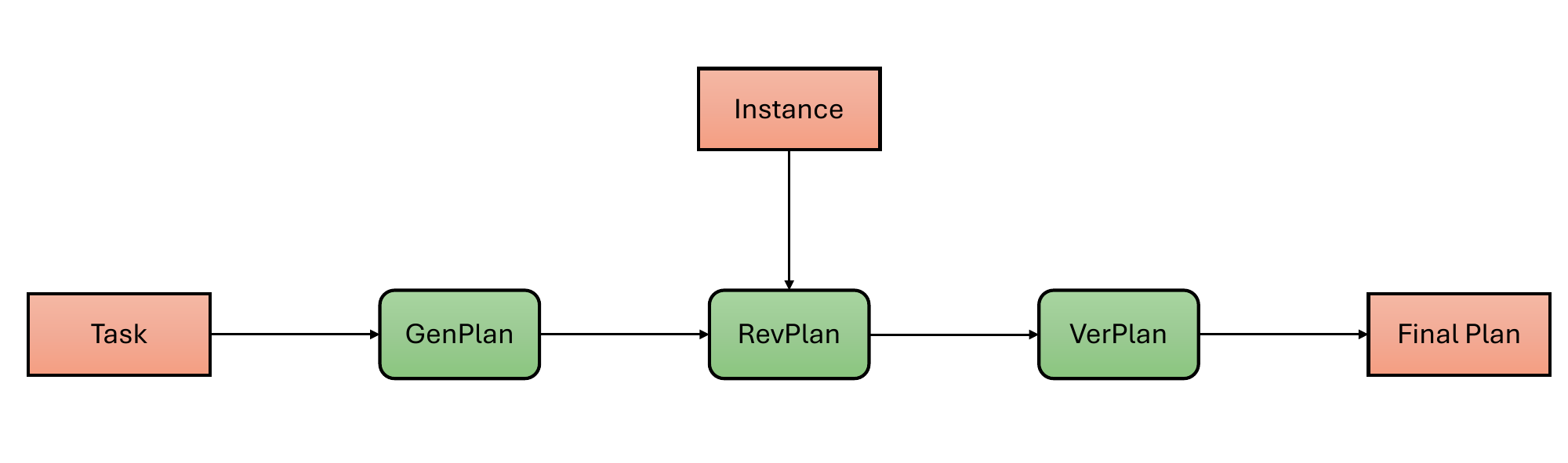}
    \caption{A description of GRASP, defining our main blocks.}
    \label{fig:GRASP_main}
\end{figure}

\section{Related Work}\label{sec:related-work}
\noindent\textbf{Autonomous Agents} Computing agents originated in 1970s rule-based expert systems like \textsc{DENDRAL} and \textsc{MYCIN} \cite{buchanan1969dendral,shortliffe1974mycin}, later evolving through decision theory and cognitive architectures into autonomous systems \cite{pearl1984probabilistic,newell1990unified,kaelbling1998planning}. Modern language model (LLM) agents act as task-directed workflows that perceive, plan, and execute within complex environments \cite{wang2025surveyhas,wang2025surveyopt,Mohammadi_2025}. Rather than acting as rigid programmatic scripts, effective agents must exhibit human-like adaptability to handle complex multi-step routines autonomously under dynamic context shifts.

\noindent\textbf{Inference-Time Algorithms} Core prompting techniques like Chain-of-Thought \cite{wei2022chain}, Self-Consistency \cite{wang2022self}, and In-Context Learning \cite{dong2024survey} enhance step-by-step reasoning but lack long-horizon planning capabilities. Search-based extensions such as Tree of Thoughts \cite{yao2023tree}, Graph of Thoughts \cite{besta2024graph}, and REBASE \cite{wu2024empirical} expand trajectory exploration to improve plan robustness, yet they lack structured constraint enforcement. While agentic frameworks attempt to separate planning from execution via incremental adjustments (ReAct \cite{yao2022react}, Pre-Act \cite{rawat2025preactmultistepplanningreasoning}) or critique loops (Reflexion \cite{shinn2023reflexion}, TextGrad \cite{yuksekgonul2024textgrad}), they remain vulnerable to conversational noise. GRASP builds on these paradigms by completely isolating generation, revision, and verification across decoupled context windows.

\noindent\textbf{Automated Planning} Multi-agent and constraint-guided frameworks mark recent advances in plan optimization. PlanGEN \cite{parmar2025plangen} combines constraint, selection, and verification agents to refine test-time search spaces. Similarly, hybrid symbolic-LLM methods like PDDL-Instruct \cite{verma2025teaching} demonstrate that guiding inference through structured rules and deterministic constraints dramatically optimizes plan validity \cite{mahdavi2024leveraging,wei2025plangenllms,cao2025large}. While advanced multi-agent orchestrators like Magentic UI \cite{mozannar2025magentic} incorporate memory and execution guards, they rely heavily on human-in-the-loop interaction for plan validation. In contrast, GRASP operates as a fully autonomous algorithm tailored specifically for high-quality, zero-shot plan generation and structural regularization.

\section{GRASP}\label{sec:design}
In this section, we introduce the GRASP (Generating, Revising, and Assessing for Strategic Planning) framework and explain the operational details of its constituent components. The framework decomposes the end-to-end planning problem into three specialized modules that are executed sequentially: (i) \emph{GenPlan}, which generates a structural blueprint of the plan strictly based on the natural language task description $\mathcal{T}$; (ii) \emph{RevPlan}, which adapts and refines this baseline plan using a localized task instance $\mathcal{I}$ by exploring distinct strategic paths (defined as heuristics in \cite{polya1945solve}); and (iii) \emph{VerPlan}, which acts as an independent evaluator to verify whether constraints are satisfied and selects the optimal plan $\mathcal{P}^*$. Each module comprises distinct computational blocks that leverage targeted prompting to keep the language model focused on the specific sub-task and operating in a reliable manner.

Architecturally, GRASP is designed to enable explicit context isolation. Standard plan generation frameworks often suffer from performance degradation because they force a model to simultaneously process global constraints, track variables, and generate local steps within the same context window. GRASP avoids this contextual overload by splitting these requirements across the three decoupled stages. We now describe each module in detail. The exact prompts, detailed configurations, and algorithmic pseudo-code for each component agent are provided in the Appendix.

\subsection{GenPlan}\label{sec:GenPlan}
The GenPlan module handles macro-level plan generation by mapping the task description $\mathcal{T}$ into a structured baseline plan $\mathcal{P}_G$ while remaining oblivious to the task instance $\mathcal{I}$. $\mathcal{P}_G$ serves as a blueprint based on which instance-specific plans will be structured. Monolithic LLMs struggle with single-shot plan generation. They produce fluent text that fails under rigorous inspection due to missing logical primitives or implicit constraint violations. This degradation occurs because standard LLMs lack explicit mechanisms to separate task parameters from runtime variables, leading to unrecoverable cascading errors. To balance the flexibility of natural language and the strict adherence to task requirements, GenPlan decomposes planning into component steps with memory managed through the knowledge base ($\mathcal{KB}$) and three specialized, prompt-engineered agents: the Constraint Agent (CA), the Guidelines Agent (GA) and the Plan Generation Agent (PGA).

Our design is conceptually grounded in formal cognitive models of human planning, specifically working memory management and self regulation. When human experts encounter dense specifications, they do not try to do detailed planning immediately. Instead, they isolate abstract environmental rules from $\mathcal{T}$, continuously reflecting and refining their mental model before applying it to concrete instances $\mathcal{I}$.

GenPlan formalizes this cognitive reflection loop. We define the $\mathcal{KB}$ as a memory store for the task, constraints, guidelines and the tentative plan. The CA, GA and PGA operate over this shared $\mathcal{KB}$ within an iterative feedback loop, balancing natural language flexibility with structural reliability. This continuous reflection process enables the framework to incrementally correct inconsistencies, uncover overlooked details, and converge towards a stable, accurate plan $\mathcal{P}_G$

\begin{figure}
    \centering
    \includegraphics[width=.8\linewidth]{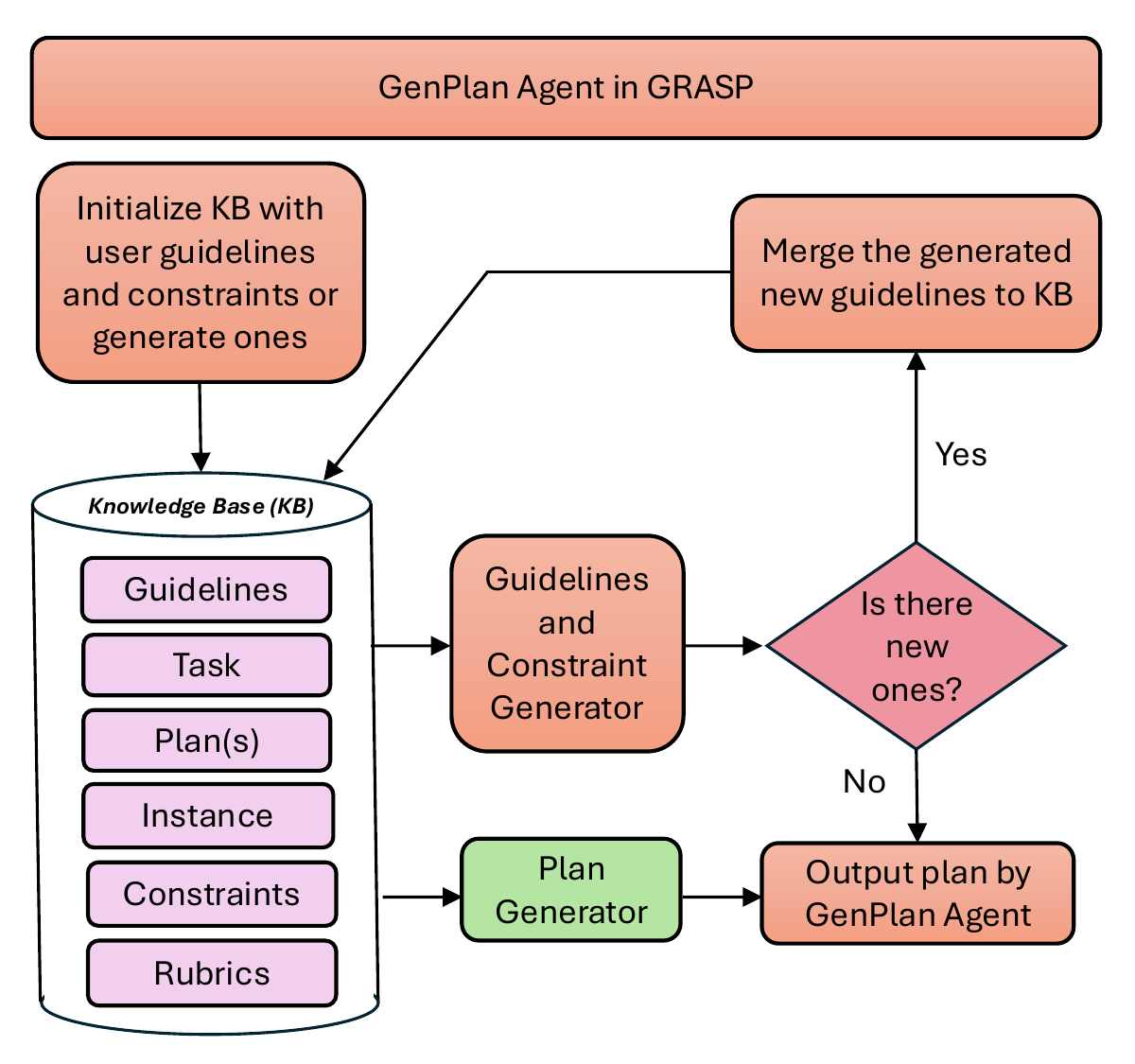}
    \caption{A block diagram of the GenPlan agent, discussed in Section~\ref{sec:GenPlan}.}
    \label{fig:GenPlan}
\end{figure}

\subsubsection{Knowledge Base}
The Knowledge Base serves as the centralized state manager for the GenPlan module. Instead of maintaining floating context variables, the framework aggregates all supporting directives such as constraints and guidelines and the tentative plan into a single structured tuple. Formally, at any iterative step $t$, we have,
\begin{align}
    \mathcal{KB}_t = \langle \mathcal{T}, \mathcal{C}_t, \mathcal{G}_t, \mathcal{P}_t \rangle
\end{align}
where $\mathcal{C}_t$ represents the set of hard constraints, $\mathcal{G}_t$ represents the set of soft guidelines, and $\mathcal{P}_t$ represents the tentative baseline plan. The $\mathcal{KB}$ handles user-specifications, which include the initial guidelines and constraints. If provided, the $\mathcal{KB}$ is initialized as,
\begin{align}
    \mathcal{KB}_0 = \langle \mathcal{T}, \mathcal{C}_{\text{init}}, \mathcal{G}_{\text{init}}, \emptyset \rangle
\end{align}
If no initial guidelines $\mathcal{G}_0$ or constraints $\mathcal{C}_0$ are provided, then they are also initialized as empty sets. The CA, GA and PGA read from the $\mathcal{KB}$ and perform updates to generate new entries or structurally modify existing entries.

\subsubsection{Guidelines Agent (GA) }
Guidelines operate as soft, non-binding regularizers or operational heuristics that are designed to optimize plan quality and coherence \cite{GEREVINI2009619}. The GA identifies new guidelines from the existing $\mathcal{KB}$ and merges them with the current guidelines. More formally,
\begin{align}
    \mathcal{G}_{t+1} = \text{MERGE}(\mathcal{G}_t, \text{GA}(\mathcal{KB}_t)). 
\end{align}
Guidelines enforce structural consistency, uniform phrasing, and boundaries across the plan. However, even if a specific guideline in $\mathcal{G}$ is violated, a candidate plan may remain logically valid. In general, enforcing guidelines minimizes generative drift, highlighting the role of guidelines as soft regularizers over the decoding space of the LLM.

\subsubsection{Constraint Agent (CA)}
In contrast to soft guidelines, constraints represent strict, necessary conditions that the plan must satisfy to be classified as executable and valid \cite{GEREVINI2009619}. The CA operates in a similar way to the GA by identifying new constraints from the existing $\mathcal{KB}$ and merges them with the current constraints. Formally,
\begin{align}
    \mathcal{C}_{t+1} = \text{MERGE}(\mathcal{C}_t, \text{CA}(\mathcal{KB}_t)). 
\end{align}
To ensure maximum plan reliability, GRASP allows users to manually enter constraints at initialization. However, even when GenPlan operates in a zero-shot fashion without user defined priors ($\mathcal{C}_0 = \phi$), the CA automatically generates a list of constraints from $\mathcal{T}$. This process ensures that the baseline plan and downstream plan synthesis follow operational boundaries. Mathematically, constraints function as hard regularizers that clip and prune logically infeasible states from the downstream plan search space.

\subsubsection{Plan Generation Agent (PGA)}
The PGA is an agent which synthesizes the tentative base plan under the structural and operational boundaries defined by $\mathcal{C}$ and $\mathcal{G}$. If user-specified guidelines and constraints are provided at $t=0$, then the PGA immediately constructs the initial plan and places it in the $\mathcal{KB}$. Otherwise, the CA and GA are invoked to build the initial constraints and guidelines. After the plan has been initialized, it is updated as follows at the $t+1^{th}$ iteration,
\begin{align}
    \mathcal{P}_{t+1} = \text{PGA}(\mathcal{KB}_t)
\end{align}

GenPlan implements the iterative optimization and symbolic reflection loop across the agents, rather than executing a single-shot forward generation pass. At each iteration $t$, the agents inspect the complete $\mathcal{KB}_t$, and generate new iterates by identifying logical mismatches or constraint violations. This process is similar to the techniques used in \cite{yuksekgonul2024textgrad, shinn2023reflexion}. The $\mathcal{KB}$ is then updated as,
\begin{align}
    \mathcal{KB}_{t+1} = \langle \mathcal{T}, \mathcal{C}_{t+1}, \mathcal{G}_{t+1}, \mathcal{P}_{t+1} \rangle.
\end{align}
This optimization loop executes continuously until an external judge block determines that the updates have converged and no further structural changes have been qualitatively observed between iteration,
\begin{align}
    \mathcal{KB}^* = \mathcal{KB}_t \quad \text{where} \quad \mathcal{KB}_{t+1} \equiv \mathcal{KB}_t
\end{align}
Through this continuous optimization loop over the $\mathcal{KB}$, the GenPlan module dynamically accumulates new constraints and guidelines, and produces the final, highly resilient baseline plan $\mathcal{P}_G \in \mathcal{KB}^*$, that strictly adheres to the hard constraints, while honoring the soft guidelines, and acts as a blueprint for downstream plan generation.

The detailed algorithm and prompts of GenPlan are given in Appendix~\ref{app:genplan}. Further, please see Appendix~\ref{app: guidelines and constraints} and \ref{app:II} for examples of guidelines, constraint and plans generated and used by GenPlan.

\subsection{RevPlan}\label{sec:Revplan}
The RevPlan module takes the macro-level baseline plan $\mathcal{P}_G$ generated by the GenPlan block, and specializes it for a specific task instance $\mathcal{I}$. In traditional software engineering, an execution instance simply supplies static data parameters to a deterministic function. In contrast, within the GRASP framework, $\mathcal{I}$ introduces dynamic runtime directives, such as edge-case exceptions, shifting operational priorities and unmodeled environmental circumstances. To satisfy these runtime directives without violating the core structural plan rules established in GenPlan, RevPlan refines the plan $\mathcal{P}_G$ by systematically introducing, modifying or removing discrete execution steps based on a multi-path exploration on the strategy space.

To prevent the model from prematurely converging on a single suboptimal decoding path, we forego a linear revision pass. Instead, as outlined in Algorithm~\ref{alg:RevPlan}, we expand the planning space into parallel, strategy-based tracks using two primary stages, strategy induction and strategy-specific plan merging.

In the strategy induction stage, we first derive a set of strategies by evaluating the task $\mathcal{T}$ and the task instance $\mathcal{I}$. We define this strategy space as,
\begin{align}
    \mathcal{S} = \{s_1, s_2, \ldots, s_K\}
\end{align}
where $K$ represents the number of distinct strategies explored. In this work, we typically consider $2 \leq K \leq 4$. This parallelization captures diverse and valid strategic paths that an LLM might otherwise discard if it is forced to commit to a single path early on.

RevPlan then initiates $K$ contextually isolated tracks, one for each strategy. Within each track, we generate a set of localized constraints $\mathcal{C}_{s_k}$, which act as rigid guardrails for the subsequent planning phase. For each isolated track, we construct a strategy-specific plan candidate $\mathcal{P}^{(k)}$ through the following three-phase pipeline.
\begin{itemize}\itemsep0em
    \item First, we leverage the ReAct framework \cite{yao2022react} to generate a strategy-specific solution trajectory $\chi_{s_k}$. The reason-observe-act framework in the ReAct solution interleaves reasoning traces with structural actions, while aligning with the strategy-specific constraints $\mathcal{C}_{s_k}$. This solution provides a valid path towards the goal of the pair of $\mathcal{T}$ and $\mathcal{I}$.
    \item We then apply an extraction function to isolate the executable steps from the verbose solution trajectory $\chi_{s_k}$,
    \begin{align}
        \mathcal{P}_{\text{specific}}^{(k)} = \text{ExtractPlan}(\chi_{s_k})
    \end{align}
    This operation isolates the structural plan from the exact solution details, yielding an instance-focused plan without the GenPlan guardrails.
    \item Finally, we reconcile this instance-specific plan with the structural blueprint provided by GenPlan,
    \begin{align}
        \mathcal{P}^{(k)} = \text{MergePlans}(\mathcal{P}_{\text{specific}}^{(k)}, \mathcal{P}_G).
    \end{align}
    This plan-merge operation preserves the structural blueprint established via $\mathcal{P}_G$, while integrating the strategy-specific logic of $\mathcal{P}_{\text{specific}}^{(k)}$.
\end{itemize}
Thus, maintaining strict contextual isolation across the $K$ tracks ensures that hallucinations and compounding errors in one trajectory do not harm other trajectories. RevPlan finally outputs a set of specialized candidate plans $\mathbb{P} = \{\mathcal{P}^{(1)}, \mathcal{P}^{(2)}, \ldots, \mathcal{P}^{(K)}\}$, representing a diverse set of valid, instance-calibrated plans.

\begin{figure}
    \centering
    \includegraphics[width=.8\linewidth]{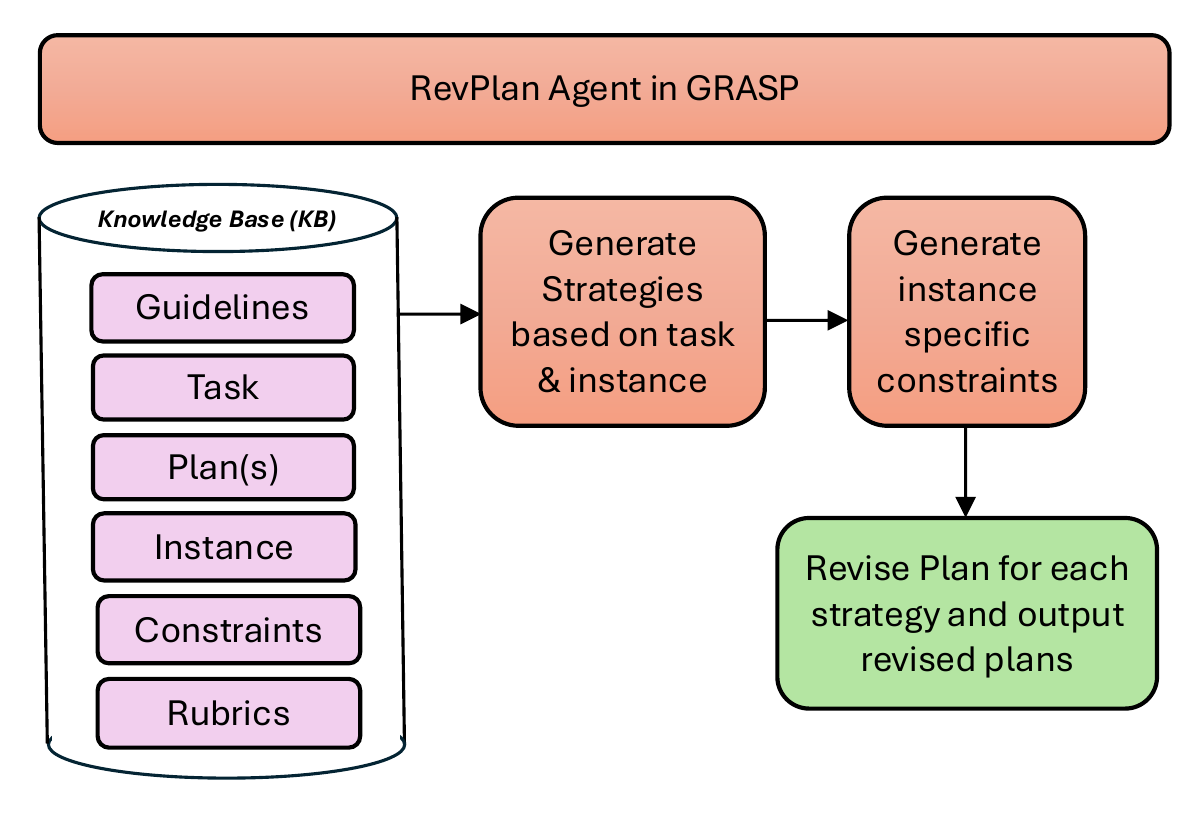}
    \caption{A block diagram of the RevPlan agent, discussed in Section~\ref{sec:Revplan}}
    \label{fig:RevPlan}
\end{figure}

In Appendix~\ref{app:revplan} and Appendix~\ref{app:II}, we present the detailed algorithm and prompts for RevPlan and examples of plan revisions based on strategies.

\begin{figure}
    \centering
    \includegraphics[width=.8\linewidth]{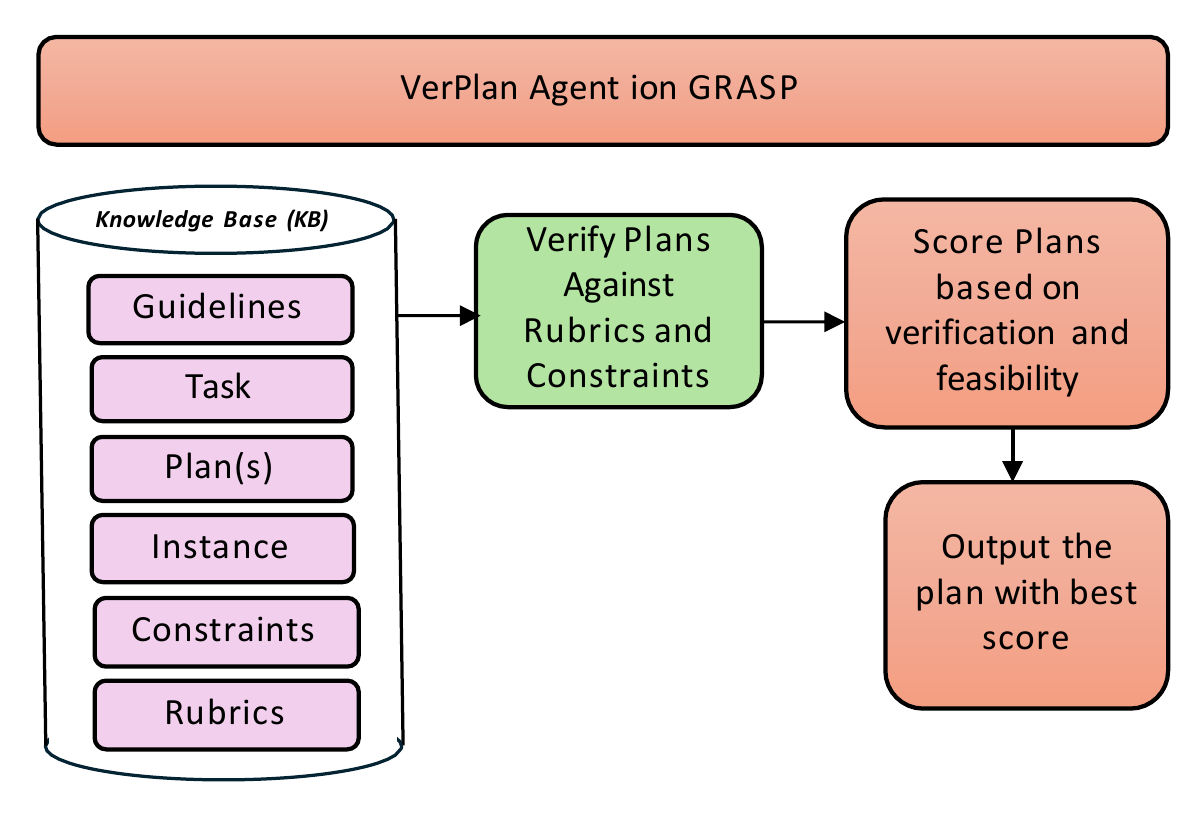}
    \caption{A block diagram of the VerPlan agent, discussed in Section~\ref{sec:Verplan}}
    \label{fig:VerPlan}
\end{figure}

\subsection{VerPlan}\label{sec:Verplan}
VerPlan is the final module of GRASP. VerPlan functions as a decoupled plan evaluator to assess the set of candidate plans $\mathbb{P}$ generated by RevPlan. We implement this verification stage because the individual strategy-based plans may still contain logical errors, omit crucial steps, or suffer from varying levels of infeasibility depending on the exact task and instance. To identify these shortcomings and filter out low-quality plans, VerPlan evaluates each candidate plan independently within an isolated context window.

Each candidate plan is assessed using an LLM, which acts as the scoring function. The LLM functions as an unbiased discriminator and scorer to provide an objective assessment of feasibility and reliability.

For each track $k$, we evaluate the candidate plan $\mathcal{P}^{(k)}$ against the task $\mathcal{T}$, instance $\mathcal{I}$, GenPlan constraints $\mathcal{C}$, strategy-specific constraints $\mathcal{C}_{s_k}$ and internal rubrics. We prompt the LLM to output a plan quality score $\Psi^{(k)} \in [0,100]$. This score reflects the plan's likelihood of success and its strict adherence to the constraints. More formally,
\begin{align}
    \Psi^{(k)} = \text{Score}(\mathcal{P}^{(k)}|\mathcal{T}, \mathcal{I}, \mathcal{C}, \mathcal{C}_{s_k}, \text{Rubrics})
\end{align}
This verification mechanism acts as a strict constraint-satisfaction and feasibility check, ensuring that plans that satisfy the constraints and are feasible are given higher scores. By scoring each trajectory independently, we prevent mistakes or noise due to one strategy path from biasing the evaluation of other candidates. Once the model has computed the evaluation scores for all $K$ candidates, we choose the plan with the highest score,
\begin{align}
    \mathcal{P}^* = arg\max_{\mathcal{P}^{(k)}\in \mathbb{P}}\Psi^{(k)}.
\end{align}
The selected plan $\mathcal{P}^*$ is chosen as the final, high-confidence output of GRASP.

\section{Evaluation}\label{sec: evaluation}

\subsection{Experimental Setup}
To evaluate the quality of plan generation with GRASP, we utilize four distinct benchmarks spanning administrative logistics, hard logic matrices, and expert-level scientific reasoning: Natural Plan Calendar Scheduling \cite{zheng2024natural}, GPQA \cite{rein2024gpqa}, SciBench Math \cite{wang2024SciBench} and ZebraLogic \cite{lin2025zebralogic}. 

\subsubsection{The Execution Platform: RunAgent} In order to guarantee a standard, objective baseline for how plans translate into actions, all generated plans across all frameworks are systematically run through RunAgent \cite{srivastava2026runagentinterpretingnaturallanguageplans}. RunAgent functions as a multi-agent execution platform that interprets natural language steps of a plan and executes them. RunAgent utilizes frontier LLMs for executing plans. RunAgent (GPT-4o \cite{hurst2024gpt}) is utilized for the Natural Plan Calendar Scheduling, Scibench Math and GPQA datasets. RunAgent (GPT-4.1-mini \cite{openai2025gpt41}) is utilized for the ZebraLogic dataset. Even though RunAgent includes many features such as constraint generation and verification, fact utilization, agentic language and running state summaries, we run a bare-bones version of RunAgent, where only the LLM or Python code execution are utilized for executing steps. Further, error correction is only invoked to overcome runtime errors. This ensures that poor plan quality is not compensated by verification protocols.

\subsubsection{Experimental Setup}
Our evaluation isolates GRASP's effectiveness in planning across two distinct experiments. More details are provided in Appendix~\ref{app: experimental details}.
\begin{itemize}\itemsep0em
    \item Single Task Planning with Programmatic execution (Experiment 1): In this experiment, a common task is described to GRASP based on the specific dataset, and each problem is provided as an instance. Each generated plan is executed with RunAgent.
    This experiment shows the utility of GRASP plans for agentic-tool-use use cases across all datasets.
    \item Robustness for Compound Task Scaling (Experiment 2): To evaluate the resilience of GRASP under dual-task objectives, we test the planners on Natural Plan Calendar Scheduling by scaling the tasks from simple tasks with single goals to tasks with two or three objectives. We turn off the Python code execution in RunAgent for this experiment, and run each step using only LLMs. 
    This experiment highlights GRASP's structural resilience under complex task loads.
\end{itemize}

\subsubsection{Evaluation Metrics and Baselines}
We report Exact Match (EM) accuracy for the Natural Plan Calendar Scheduling dataset, and standard accuracy percentages for the SciBench Math, GPQA and ZebraLogic datasets. Equivalence between RunAgent outputs and gold answers is judged using an LLM judge block, running with GPT-5 \cite{openai2025gpt5}. Manual evaluation of 50 random instances per dataset classified by GPT-5 demonstrates full agreement with human judgment. For Experiment 1, we benchmark GRASP against three types of baselines:
\begin{itemize}\itemsep0em
    \item Direct LLM execution: Single-shot problem solving by baseline LLMs.
    \item Baseline LLM Planner: Single-shot plan generated by LLM based on task and instance with structural guardrails in the prompt, and executed with RunAgent.
    \item State-of-the-art Methods: Advanced algorithms including PlanGEN \cite{parmar2025plangen} (all four variants), Tree-of-thoughts (ToT) \cite{yao2023tree} and Best-of-N (BoN) \cite{Gui24bonbon} are used to generate plans, which are then executed by RunAgent.
\end{itemize}
To rigorously examine whether GRASP can outperform frontier reasoning models in plan generation for dual-task objectives, we introduce GPT-5-mini (medium reasoning effort) \cite{openai2025gpt5} as a baseline direct planner alongside GPT-4o and GPT-4o-mini for comparison in Experiment 2.

\begin{table}[t]
    \centering
    \setlength{\tabcolsep}{4pt}
    \small
    \begin{tabular}{lc}
    \hline
    \textbf{Method}  &  \textbf{EM Acc.} ($\%$)\\
    \hline
    GPT-4o-mini baseline & 41.9\\
    GPT-4o-mini Planner & 54.3\\
    GRASP(GPT-4o-mini) planner & 74.3\\
    GPT-4o baseline & 58.3\\
    GPT-4o Planner & 63\\
    GRASP(GPT-4o) planner & 75.4\\
    \hline
    \end{tabular}
    \caption{Results for Natural Plan Calendar Scheduling}
    \label{tab:calendar results}
\end{table}

\begin{table}[t]
    \centering
    \setlength{\tabcolsep}{4pt}
    \small
    \begin{tabular}{lc}
    \hline
    \textbf{Method}  &  \textbf{Accuracy} ($\%$)\\
    \hline
    GPT-4o baseline & 29.5\\
    GPT-4o Planner & 30.6\\
    GRASP (GPT-4o) Planner & 61.4\\
    \hline
    \end{tabular}
    \caption{{Results for the ZebraLogic Dataset}}
    \label{tab: zebralogic results}
\end{table}

\begin{table}[t]
    \centering
    \setlength{\tabcolsep}{4pt}
    \small
    \begin{tabular}{lccc}
    \hline
    \textbf{Method}  &  \textbf{Stat} & \textbf{Calc} & \textbf{Diff}\\
    \hline
    GPT-4o-mini baseline & 73.61 & 65.85 & 34\\
    GPT-4o-mini Planner  & 70.83 & 73.17 & 28\\
    GRASP(GPT-4o-mini) planner & 81.94 & 73.17 & 50\\
    GPT-4o baseline & 77.78 & 70.73 & 50\\
    GPT-4o Planner & 70.83 & 73.17 & 54\\
    GRASP(GPT-4o) planner & 80.56 & 78.05 & 62\\
    \hline
    \end{tabular}
    \caption{{Accuracy Percentage for the SciBench Dataset}}
    \label{tab:SciBench results}
\end{table}

\begin{table}[t]
    \centering
    \setlength{\tabcolsep}{4pt}
    \small
    \begin{tabular}{lc}
    \hline
    \textbf{Method}  &  \textbf{Accuracy} ($\%$)\\
    \hline
    GRASP (GPT-4o, 2 strat.) planner & 47.54\\
    GPT-4o baseline & 47.99\\
    GRASP (GPT-4o, 4 strat.) planner & 48.23\\
    \hline
    \end{tabular}
    \caption{{Results for the GPQA Dataset}}
    \label{tab:gpqa results}
\end{table}

\begin{table*}[htbp]
\centering
\setlength{\tabcolsep}{4pt}
\small
\begin{tabular}{lccccc}
\toprule
\textbf{Method} & \textbf{Calendar Scheduling} & \textbf{SciBench Stat} & \textbf{SciBench Calc} & \textbf{SciBench Diff} & \textbf{ZebraLogic} \\
\midrule
PlanGEN (Mix)    & 52.3 & 72.22 & 73.17 & 50.0 & 31.5 \\
PlanGEN (ToT)    & 55.7 & 50.00 & 40.48 & 26.0 & 10.0 \\
PlanGEN (Rebase) & 49.4 & 77.78 & 64.29 & 34.0 & 39.2 \\
PlanGEN (BoN)    & 72.3 & 73.16 & 66.67 & 48.0 & 41.6 \\
BoN              & 60.4 & 77.78 & 70.73 & 34.0 & 41.0 \\
ToT              & 52.2 & 58.33 & 48.78 & 40.0 & 13.0 \\
\hline
GRASP (Ours) & 74.3 & 81.94 & 73.17 & 50.0 & 58.0 \\
\bottomrule
\end{tabular}
\caption{Performance Comparison of PlanGEN Variants, Baselines, and GRASP Across Benchmarks}
\label{tab:performance_comparison}
\end{table*}

\begin{table}[ht!]
    \centering
    \setlength{\tabcolsep}{4pt}
    \small
    \begin{tabular}{lccc}
    \hline
    \textbf{Method}  &  \textbf{Input} & \textbf{Output} & \textbf{NC}\\
    \hline
    GPT-4o Baseline & 262 & 352 & 1\\
    GPT-4o Planner & 374 & 418 & 1.2\\
    GRASP(GPT-4o, $1$ strat.) & 4703 & 1316 & 6\\
    GRASP(GPT-4o, $2$ strat.) & 8497 & 2400 & 10.8\\
    GRASP(GPT-4o, $3$ strat.) & 12664 & 3741 & 16.5\\
    GRASP(GPT-4o, $4$ strat.) & 16455 & 4753 & 21.2\\
    GRASP(GPT-4o-mini, $3$ strat.) & 13683 & 4711 & 1.16\\
    \hline
    \end{tabular}
    \caption{{Average token count and normalized cost (NC) per instance for Natural Plan Calendar Scheduling}}
    \label{tab:token and cost results}
\end{table}

\begin{table}[h]
    \centering
    \small
    \setlength{\tabcolsep}{3pt}
    \begin{tabular}{lcc}
    \hline
    \textbf{Method} & \textbf{Dual Tasks} & \textbf{Triple Tasks} \\
    \hline
    GPT-4o-mini Planner & 24.5 & 21.2 \\
    GPT-4o Planner & 30.5 & 29.1 \\
    GPT-5-mini Planner & 26.7 & 25.5 \\
    \hline
    \textbf{GRASP(GPT-4o-mini)} & \textbf{41.2} & 28.1 \\
    \textbf{GRASP(GPT-4o )} & \textbf{45.7} & \textbf{46.6} \\
    \hline
    \end{tabular}
    \caption{Performance under multi-task scaling.}
    \label{tab:multi_task_robustness}
\end{table}

\begin{table}[h]
    \centering
    \small
    \setlength{\tabcolsep}{10pt}
    \begin{tabular}{clc}
    \hline
    \textbf{Setting} & \textbf{Configuration} & \textbf{Accuracy (\%)} \\
    \hline
    1 & Direct Planner & 30.5 \\
    2 & RevPlan Only & 10.9 \\
    3 & VerPlan Only & 30.7 \\
    4 & GenPlan Only & 40.8 \\
    5 & RevPlan + VerPlan & 12.5 \\
    6 & GenPlan + RevPlan & 42.5 \\
    7 & \textbf{GRASP} & \textbf{45.7} \\
    \hline
    \end{tabular}
    \caption{Ablation matrix across GRASP modules under dual-task mode.}
    \label{tab:7_way_ablation}
\end{table}

\subsection{Main Results}\label{sec: main results}
We summarize and discuss our main results in this section. Statistical analysis, plan efficiency analysis and impacts of information leakage are investigated in Appendix~\ref{app: statistical validation}.

\subsubsection{Single-Task Planning with Programmatic Execution}
\noindent\textbf{Performance on Natural Plan Calendar Scheduling} The results for this dataset are summarized in Table~\ref{tab:calendar results}. We observe that GRASP outperforms LLM planners by $12.4\%$ for GPT-4o and by $20\%$ for GPT-4o-mini. Surprisingly, GRASP(GPT-4o-mini) outperforms the GPT-4o planner by $11.3\%$.

\noindent\textbf{Performance on ZebraLogic} The results are summarized in Table~\ref{tab: zebralogic results}. In this dataset, GRASP(GPT-4o) outperforms the GPT-4o planner by $30.8\%$.

\noindent\textbf{Performance on SciBench Math} The results are summarized in Table~\ref{tab:SciBench results}. GRASP(GPT-4o) outperforms the GPT-4o planner on all subsets. GRASP(GPT-4o-mini) outperforms the GPT-4o-mini planner on the Stat and Diff subsets, but equals on the Calc subset. On the Stat subset, GRASP(GPT-4o-mini) outperforms other methods.

\noindent\textbf{Performance on GPQA} The results for this dataset are summarized in Table~\ref{tab:gpqa results}. The results are averaged over $3$ runs, and we find that GRASP does not perform better than either the GPT-4o baseline or the planner in a statistically significant way.

\noindent\textbf{Comparison against State-of-the-art} The state-of-the-art algorithms and GRASP perform planning using GPT-4o-mini, and each plan is executed using RunAgent. The results are summarized in Table~\ref{tab:performance_comparison}. We observe that GRASP outperforms all other methods on Natural Plan Calendar Scheduling, SciBench Stat, and ZebraLogic dataset, but PlanGEN equals GRASP on SciBench Calc and Diff.

\noindent\textbf{Ablation Studies} We conduct ablation studies on Natural Plan Calendar Scheduling. Varying the number of strategies in RevPlan (1–4) yields accuracies of $73.0\%$, $74.1\%$, $75.4\%$, and $71.2\%$, respectively, indicating a performance sweet spot at three strategies and highlighting the impact of strategy initialization on plan generation.
By using 3 strategies in RevPlan and removing constraint and guideline generation in GenPlan the accuracy drops from $75.4\%$ to $58.5\%$.

\noindent\textbf{Token and Cost Comparisons} We summarize the average input and output token counts per instance for Natural Plan Calendar Scheduling in Table~\ref{tab:token and cost results}. Using the current GPT token prices (see Appendix~\ref{app: token prices}), we compute the cost and report the normalized cost versus baseline GPT-4o model in Table~\ref{tab:token and cost results}. GRASP (GPT-4o) increases cost by $\sim13.5$x while improving accuracy by $12.4\%$ over the GPT-4o planner. GRASP (GPT-4o-mini) with three strategies achieves an $11.3\%$ accuracy gain at $\sim0.95$x the cost. Further discussion of token usage for Table~\ref{tab:performance_comparison} is given in Appendix~\ref{app: token usage}.

\subsubsection{Robustness for Compound Task Scaling}
\noindent\textbf{Performance under Multi-Task Scaling} The results for multi-task scaling 
are summarized in Table~\ref{tab:multi_task_robustness}. While all planners achieve parity on isolated single tasks (hovering between $44.5\%$ and $45.8\%$), expanding the workspace to interleaved multi-task streams causes direct planners to collapse due to context pollution. Moving from dual to triple tasks, the direct GPT-4o planner decays from $30.5\%$ to $29.1\%$, and the GPT-4o-mini planner drops from $24.5\%$ to $21.2\%$. Conversely, GRASP(GPT-4o) maintains performance, achieving $45.7\%$ on dual tasks and rising to $46.6\%$ on triple tasks, outperforming its direct planner counterpart by $17.5\%$. GRASP also enables smaller LLMs to excel on dual tasks, driving GRASP(GPT-4o-mini) to $41.2\%$ (a $16.7\%$ absolute gain over its direct baseline).

\noindent\textbf{Comparison Against Reasoning Models} We evaluate direct planners driven by GPT-5-mini, a frontier reasoning model. Despite this optimization, the direct GPT-5-mini planner succumbs to multi-task context degradation, scoring only $26.7\%$ on dual tasks and dropping to $25.5\%$ on triple tasks. Both GRASP configurations comprehensively outperform this frontier reasoning model. Crucially, GRASP(GPT-4o-mini) achieves $41.2\%$ on dual tasks, outperforming the direct GPT-5-mini planner by a $14.5\%$ absolute margin.

\noindent\textbf{Ablation Studies} We execute an ablation study under the dual-task setup, summarized in Table~\ref{tab:7_way_ablation}. Isolating single modules shows that GenPlan Only yields $40.8\%$ accuracy, proving the vital anchoring role of guidelines and constraints. In contrast, activating local strategy exploration without structural boundaries (RevPlan Only or RevPlan + VerPlan) causes a catastrophic drop to $10.9\%$ and $12.5\%$, respectively—performing substantially worse than VerPlan Only ($30.7\%$). The highest overall performance is unlocked strictly when all modules interact in the full GRASP architecture ($45.7\%$).

\noindent\textbf{Discussion} The ablation study reveals a profound modular interdependence. VerPlan Only acts as a passive filter over standard direct generations, yielding baseline levels. However, executing local updates (RevPlan) without the invariant structural guardrails of GenPlan triggers unconstrained semantic drift and plan hallucinations. This creates a set of low quality plans, catching the downstream VerPlan module in a classic "garbage-in, garbage-out" trap that drops performance to $12.5\%$. This confirms that local strategy optimization requires global regularization. Additionally, our triple-task results uncover a clear capacity boundary condition for smaller models. While GRASP(GPT-4o-mini) performs robustly on dual tasks ($41.2\%$), it collapses to $28.1\%$ on triple tasks. This demonstrates that under extreme multi-task workloads, smaller LLMs lack the raw parameter capacity to successfully manage complex instructions without suffering local attention drift, establishing a hardware-bound capability threshold.

\section{Conclusion}\label{sec: conclusion}
In this paper, we introduced GRASP, an autonomous framework that neutralizes context degradation and plan hallucinations in complex tasks. By decoupling plan generation, strategy exploration, and validation across context-isolated agents, our architecture successfully eliminates compounding error cascades. Our evaluations prove that explicit structural guardrails enable standard LLMs to systematically outperform the unguided internal reasoning chains of frontier reasoning models.

\section*{Limitations}\label{sec: limitations}
\noindent\textbf{Latency Considerations} While our framework prioritizes execution reliability, the iterative generation, local refinement, and systematic verification of multiple alternative strategies inherently incurs higher runtime latency compared to single-pass, monolithic generation. We consider this additional computational overhead a necessary trade-off to ensure structural robustness and eliminate compounding error loops in highly complex tasks. However, this latency characteristic may limit the framework's suitability for real-time, interactive applications where immediate response times are critical, defining a clear trajectory for future work to explore parallelized generation pipelines or early-exit optimization heuristics to reduce inference time.

\noindent\textbf{Orthogonal to Model-Level Reasoning} A fundamental characteristic of GRASP is that it operates strictly as an inference-time plan regularizer rather than a mechanism for improving the intrinsic, low-level reasoning capabilities of the underlying language model. If a backbone possesses severe baseline domain ignorance or lacks the foundational logic to understand a given task space, our framework cannot bridge that cognitive deficit. GRASP is explicitly designed to optimize plan execution safety by decoupling context and enforcing macro-constraints, allowing models to fully express their latent capabilities without attention fatigue. It does not train or structurally alter the underlying model's multi-stage internal reasoning chains. This limitation is specifically highlighted in our experiments. We observe that GRASP does not outperform the direct planner on the GPQA dataset, the SciBench Diff and the SciBench Calc datasets. Problems in all three datasets rely not just on building a good plan, but on improving the reasoning capability of the underlying LLM itself.

\noindent \textbf{Task-Dependent Strategy Tuning} A key limitation of the current framework is that the optimal number of generated strategy-based plans behaves as a task-dependent hyperparameter rather than a fixed, universal value. For instance, in our Calendar Scheduling evaluations, system performance peaks sharply at a sweet spot of three candidate strategies; while fewer strategies limit the explored search space, generating more introduces excessive diversity that can mislead the VerPlan discriminator toward sub-optimal choices. Consequently, deploying GRASP across highly heterogeneous domains currently requires manual tuning to balance strategic exploration with downstream execution reliability, highlighting a valuable avenue for automated, dynamically adjusting number of strategies in future iterations.

\section*{Ethical Considerations}
\paragraph{OpenAI Policy Compliance:} The use of GPT-4o, GPT-4o-mini and GPT-5-mini throughout this work adheres to the OpenAI developer terms, data governance guidelines and safety usage policies.

\paragraph{Dual-Use and Misuse Risks:} GRASP increases the structural reliability, logical consistency and execution fidelity of complex multi-step plans, and could be potentially adapted by malicious actors to optimize harmful plans. We note that GRASP functions strictly as a plan optimization framework, and lacks physical agency, real-world permissions, or independent tool-use capabilities. We strongly advocate for the enforcement of strict verification guardrails and human-in-the-loop validation protocols in any production environments that deploy planners such as GRASP.

\paragraph{Bias and Safety Alignment:} GRASP fundamentally relies on the internal semantic representations and latent knowledge space of its underlying base LLM. While it enforces structural correctness, constraint verification, and soft guideline regularization, it doesn't natively correct or filter out embedded historical, cultural, or demographic biases hidden in the base model's weights. If user-provided parameters or instructions introduce biased or problematic premises, running GRASP may produce plans that are structurally sound and optimized, but are ethically compromised or socially harmful.

\paragraph{Privacy and Data Governance:} Our work utilizes standard and publicly available benchmarks, but executes real-world planning tasks such as calendar scheduling involves handling highly sensitive user data. Multi-agent pipelines risk context-bleed or unauthorized data persistence. Care should be taken to prevent sensitive parameters from leaking into public training loops.

\paragraph{Environmental Impact and Computational Overhead:} Iterative multi-agent frameworks lead to compounding computational costs and environmental footprints, if not engineered with care. Poorly optimized multi-agent setups regularly trap LLMs in verbose, unconstrained self-reflection loops that use huge number of tokens, incurring substantial environmental costs for negligible accuracy returns. GRASP mitigates this environmental concern through its design. As shown in Sec.~\ref{sec: main results}, GRASP provides performance gains while limiting token costs. Moreover GRASP's token usage is highly competitive when compared to other baselines, as shown in Table~\ref{tab:token_efficiency}. This results in a reasonable overhead for the environmental impact and computational costs of GRASP for the performance gains provided.

\bibliography{refs}

\appendix

\begin{table*}[t]
\centering
\small
\setlength{\tabcolsep}{4pt}
\begin{tabular}{lccccc}
\hline
\textbf{Method} & \textbf{Calendar Scheduling} & \textbf{SciBench Stat} & \textbf{SciBench Calc} & \textbf{SciBench Diff} & \textbf{ZebraLogic} \\
\hline
PlanGEN (Mix)    & 15,823 / 7,095   & 13,344 / 6,375   & 12,783 / 6,196   & 13,796 / 6,683   & 17,566 / 7,918   \\
PlanGEN (ToT)    & 144,970 / 30,709 & 85,338 / 25,485  & 129,245 / 34,948 & 117,901 / 34,309 & 173,987 / 56,963 \\
PlanGEN (Rebase) & 9,318 / 4,239    & 7,072 / 31,35    & 6,627 / 3,001    & 7,684 / 3,489    & 11,973 / 5,166   \\
PlanGEN (BoN)    & 16,776 / 6,064   & 13,804 / 5,865   & 13,297 / 5,706   & 14,624 / 6,246   & 20,297 / 7,686   \\
Best of $N$      & 5,032 / 905      & 2,871 / 822      & 3,884 / 1,078    & 3,935 / 1,158    & 5,072 / 1,275    \\
Tree-of-Thoughts & 342,951 / 14,316 & 627,831 / 48,529 & 456,324 / 36,949 & 539,018 / 43,308 & 743,532 / 33,621 \\
\hline
\textbf{GRASP (Ours)} & \textbf{13,683 / 4,711} & \textbf{10,754 / 4,652} & \textbf{14,878 / 6,854} & \textbf{15,474 / 7,432} & \textbf{14,026 / 4,936} \\
\hline
\end{tabular}
\caption{Comprehensive token cost matrix across target benchmarks. Metrics are presented as \textit{Total Input Tokens} / \textit{Total Output Tokens}.}
\label{tab:token_efficiency}
\end{table*}

\section{Future Work}\label{sec: future-work}
We plan to extend our plan-generation framework by incorporating a human-in-the-loop component. In the current design, human input can be injected alongside the task and instance specification—for example, by supplying additional guidelines, facts, or constraints beyond those automatically inferred by our algorithm. In practice, however, task specifications may contain ambiguous or incomplete information. To address this, we plan to introduce an interactive chat interface that automatically formulates clarification queries to the user. While generating questions about missing or unclear information is relatively straightforward, using the same interface to solicit high-level guidance for plan construction (e.g., decomposition strategies or expert heuristics) is significantly more challenging.

A second extension involves enabling runtime plan modification by examining the output of each step and adjusting the plan dynamically. This includes verifying the correctness and relevance of intermediate results and re-executing steps when necessary by feeding verification feedback back into the LLM. In more complex cases, the system may need to insert new steps or revise existing ones to ensure proper plan execution. The evaluation of step-level outputs can leverage task-specific rubrics, and the interactive chat window also allows the system to request such rubrics from the human when dealing with critical stages of the plan.

\section{Token Prices}\label{app: token prices}
We list the prices for input and output tokens for GPT-4o, GPT-4o-mini  and GPT-5-mini in Table~\ref{tab:token prices}. This cost is in dollars, and per million tokens.

\begin{table}[h]
    \centering
    \small
    \begin{tabular}{ccc}
    \hline
    Model & Input Price$(\$)$ & Output Price$(\$)$\\
    \hline
    GPT-4o & 2.5 & 10\\
    GPT-4o-mini & 0.15 & 0.6\\
    GPT-5-mini & 0.25 & 2\\
    \hline
    \end{tabular}
    \caption{Token Prices for the models}
    \label{tab:token prices}
\end{table}

\section{Token Usage}\label{app: token usage}
The token usage for state-of-the-art baseline methods and GRASP is given in Table~\ref{tab:token_efficiency}. We observe that GRASP's token usage is competitive with other algorithms such as PlanGEN(Mix) and PlanGEN(BoN). PlanGEN(Rebase) and Best of $N$ algorithms consume less tokens than GRASP, but also perform worse than GRASP across datasets. PlanGEN(ToT) and ToT consume a large number of tokens, and also perform worse than GRASP. On the Natural Plan Calendar Scheduling dataset, PlanGEN (BoN) performs very well, following the trend shown in \cite{parmar2025plangen} for the Natural Plan datasets. However, it performs worse than GRASP, while consuming $22\%$ extra tokens. PlanGEN models perform at the same level as GRASP on the SciBench Calc and SciBench Diff datasets. This solidifies the discussion in Section~\ref{sec: limitations} that GRASP does not improve the reasoning power of the underlying LLM powering the framework.

\section{Experimental Details}\label{app: experimental details}
To ensure deterministic behavior, we set the temperature of all models to $0$ while running GRASP and RunAgent. For baselines ToT and PlanGEN, we used the temperature and other parameters suggested in the original works. We utilized gpt-4o-2024-11-20 as the gpt-4o version, gpt-4o-mini-2024-07-18 for gpt-4o-mini, gpt-4.1-mini-2025-04-14 for gpt-4.1-mini, gpt-5-mini-2025-08-07 for gpt-5-mini and gpt-5-2025-08-07 for gpt-5. Due to the high costs for each run, we only perform one run for each dataset for each experiment and model. The only exceptions are in the case of the GPQA dataset, where we perform $3$ runs, and the ablation study for the second experiment, where we run GRASP and GenPlan+RevPlan $3$ times to obtain statistical results.

\subsection{Single-Task Planning with Programmatic Execution}

We run the direct baseline LLMs with temperature $0$, by giving the direct problem from each dataset without the task or any other text. We run the direct LLM planners with the prompt provided in Appendix~\ref{app: gpt-4o plan generation prompt} under the prompt for Experiment 1. 

\paragraph{BoN, Tree-of-Thought (ToT) and PlanGEN baselines.}
We evaluate standalone BoN planners (Four direct plans are generated using the prompt for Experiment 1 in Appendix~\ref{app: gpt-4o plan generation prompt}, and the plan that scores the highest accuracy with VerPlan is executed), Tree-of-Thought (directly implemented using the code given in the Github Repository) planners and four PlanGEN (directly implemented using the code given in the Github Repository) algorithm variants, each followed by execution with \textsc{RunAgent}(Python). BoN is run with $n_{\mathrm{plans}}=4$, and VerPlan is used to choose the best plan. The temperature is set to $0$ in this baseline. The standalone ToT search uses a Princeton-style propose-value-greedy procedure with at most 15 steps, branching factor 4, beam width 1, and 4 independent value samples per candidate; generation and evaluation use temperature $1.0$ (proposal \texttt{max\_tokens}$=300$, value \texttt{max\_tokens}$=800$). Partial plans are scored by averaging mapped labels (\textit{sure}$=20$, \textit{maybe}$=1$, \textit{impossible}$=0.001$). Unless noted otherwise, the planner is \texttt{gpt-4o-mini}, the executor is \texttt{gpt-4o}, and answers are judged by \texttt{gpt-5} at temperature $1.0$.

PlanGEN runs share generation temperature $0.7$, \texttt{max\_tokens}$=1024$, judge \texttt{gpt-5}, and (for the plan-and-execute pipeline) planner \texttt{gpt-4o-mini} with executor \texttt{RunAgent(gpt-4o)}. The four variants are: \textbf{(i) Mixture of Algos.}: full constraint-verification-selection with 3 candidate solutions; \textbf{(ii) best-of-$n$}: $n_{\mathrm{plans}}=4$, diverse sampling; \textbf{(iii) tree-of-thought}: branching factor 3, max depth 20, beam width 2; and \textbf{(iv) REBASE}: 5 refinement iterations with improvement threshold $0.1$.

\subsection{Robustness for Compound Task Scaling}
Standard calendar-scheduling instances require finding one feasible meeting time. Dual and triple tasks extend this into composite objectives: the model must solve the original scheduling problem and one or two auxiliary consultant-calendar subproblems, returning a strictly formatted multi-line answer. A \textbf{dual task} has two sub-objectives, with output on exactly two newline-separated lines (no extra text). This prompt is given in Table~\ref{tab: dual task objective}. A \textbf{triple task} adds a third sub-objective (three newline-separated lines must be output). This prompt is given in Table~\ref{tab: triple task objective}. 

Each instance is composed of two parts. The first part contains the original instance of the Natural Plan Calendar Scheduling zero-shot prompt, and the second part contains the schedule of Consultant D, shown in Table~\ref{tab: consultant d schedule}, or the schedules of both Consultant D and Consultant E (shown in Table~\ref{tab: consultant e schedule}). In a similar way, the first part of the gold answers corresponds to the gold answer of the associated calendar scheduling instance, and the second part of the gold answer corresponds to the fixed answer for the latest free slot of Consultant D's schedule for the dual task, or the latest free slot in both Consultant D and Consultant E's schedule, for the triple task instruction.

The first task measures performance on standard calendar scheduling instances. Therefore, this part of the prompt varies per instance. The second part of the prompt is fixed for every instance, and measures whether the agent can parse dense calendars, apply working-hour bounds, identify 30-minute gaps, and select the latest slot, not merely a free slot. The intent of the second part of the prompt is not to be a challenging problem. The auxiliary tasks are designed to be very simple, so that the individual LLMs can solve them easily. However, when the auxiliary tasks are combined with calendar scheduling tasks, the plan accuracy goes down. 

We use GRASP with 4 strategies for both GPT-4o and GPT-4o-mini, and use the prompt in Appendix~\ref{app: gpt-4o plan generation prompt} (under Experiment 2) for the direct LLM planners.

\paragraph{Ablation Setup} Below, we describe the exact setup of each method in the ablation study in Table~\ref{tab:7_way_ablation}.
\begin{itemize}
    \item \textbf{Direct Planner}: A plan is directly generated using the prompt provided in Appendix~\ref{app: gpt-4o plan generation prompt} for Experiment 2. This new prompt ensures low plan leakage by GPT-5-mini.
    \item \textbf{GenPlan Only}: A plan is generated using the GenPlan module, and run for all instances.
    \item \textbf{RevPlan Only}: Four strategy-based plans are generated, and one of them is randomly chosen and executed.
    \item \textbf{VerPlan Only}: Four plans are generated using the  Direct Planner prompt, and VerPlan chooses the best one by scoring each plan. The plan with the highest score is executed.
    \item \textbf{GenPlan+RevPlan}: First a plan is generated by GenPlan based on the task, then four plans are generated by RevPlan based on strategies, and normalized by GenPlan's output. Then a plan is randomly chosen for execution.
    \item \textbf{RevPlan+VerPlan}: Four strategy-based plans are generated by RevPlan, and then scored by VerPlan. The best plan is chosen for execution.
    \item \textbf{GRASP}: The full framework is used to generate a good plan.
\end{itemize}

\begin{table}[htbp]
\centering
\begin{tcolorbox}[colback=gray!10, colframe=black, title=Dual Task Prompt]
\footnotesize Composite Task Instructions:\\
Your objective consists of two distinct scheduling tasks. You must output your final response on exactly two lines "$\backslash$n" separated, with no additional conversational text.\\
Line 1 (Task 1): Output the final schedule for the primary group meeting in the format: Day, HH:MM AM/PM - HH:MM AM/PM\\
Line 2 (Task 2): Analyze the 5-day schedule for External Consultant D. Find the absolute latest 30-minute free interval available in their standard working week (Monday through Friday, 9:00 AM to 5:00 PM) and output only the day and start time in the format: Day, HH:MM AM/PM
\end{tcolorbox}
\caption{}
\label{tab: dual task objective}
\end{table}

\begin{table}[htbp]
\centering
\begin{tcolorbox}[colback=gray!10, colframe=black, title=Triple Task Prompt]
\footnotesize Composite Task Instructions:\\
Your objective consists of three distinct scheduling tasks. You must output your final response on exactly three lines "$\backslash$n" separated, with no additional conversational text.\\
Line 1 (Task 1): Output the final schedule for the primary group meeting in the format: Day, HH:MM AM/PM - HH:MM AM/PM\\
Line 2 (Task 2): Analyze the 5-day schedule for External Consultant D. Find the absolute latest 30-minute free interval available in their standard working week (Monday through Friday, 9:00 AM to 5:00 PM) and output only the day and start time in the format: Day, HH:MM AM/PM\\
Line 3 (Task 3): Analyze the 5-day schedule for External Consultant E. Find the absolute latest 30-minute free interval available in their standard working week (Monday through Friday, 9:00 AM to 5:00 PM) and output only the day and start time in the format: Day, HH:MM AM/PM
\end{tcolorbox}
\caption{}
\label{tab: triple task objective}
\end{table}

\begin{table}[]
\centering
\begin{tcolorbox}[colback=gray!10, colframe=black, title=Consultant D Schedule Prompt]
\footnotesize External Consultant D's Calendar Log (Standard Hours: 9:00 AM - 5:00 PM):\\

Monday\\
* 09:00 AM - 11:30 AM: Locked Core Architecture Review\\
* 11:30 AM - 12:00 PM: [No events scheduled]\\
* 12:00 PM - 03:00 PM: Mandatory Security Infrastructure Patching\\
* 03:00 PM - 05:00 PM: HR Policy Alignment \& Compliance Audit\\

Tuesday\\
* 09:00 AM - 01:30 PM: External Vendor Negotiations (Do Not Disturb)\\
* 01:30 PM - 03:30 PM: Urgent Executive Escalation Sync\\
* 03:30 PM - 05:00 PM: Critical Server Migration Oversight\\

Wednesday\\
* 09:00 AM - 12:00 PM: Legal \& Privacy Framework Review\\
* 12:00 PM - 01:00 PM: Stakeholder Alignment Working Lunch\\
* 01:00 PM - 04:00 PM: Quarterly Budget Outlining Workshop\\
* 04:00 PM - 04:30 PM: [No events scheduled]\\
* 04:30 PM - 05:00 PM: Hard Stop - Medical Appointment\\

Thursday\\
* 09:00 AM - 05:00 PM: All-Day Offsite Leadership Bootcamp (Fully Unavailable)\\

Friday\\
* 09:00 AM - 12:30 PM: Production Deployment Readiness Check\\
* 12:30 PM - 03:30 PM: Cross-Departmental Post-Mortem Analysis\\
* 03:30 PM - 04:00 PM: Client Escalation Debrief\\
* 04:00 PM - 04:30 PM: [No events scheduled]\
* 04:30 PM - 05:00 PM: Mandatory Weekly Sign-off \& System Freeze
\end{tcolorbox}
\caption{}
\label{tab: consultant d schedule}
\end{table}

\begin{table}[]
\centering
\begin{tcolorbox}[colback=gray!10, colframe=black, title=Consultant E Schedule Prompt]
\footnotesize External Consultant E Calendar Log (Standard Hours: 9:00 AM - 5:00 PM):\\

Monday\\
* 09:00 AM - 01:00 PM: Cross-Regional Architecture Synchronization\\
* 01:00 PM - 04:30 PM: High-Severity Outage Post-Mortem \& Remediation\\
* 04:30 PM - 05:00 PM: Executive Steering Committee Briefing\\

Tuesday\\
* 09:00 AM - 10:30 AM: Bi-Annual Infrastructure Budget Defend\\
* 10:30 AM - 11:00 AM: [No events scheduled]\\
* 11:00 AM - 02:00 PM: Critical Cloud Migration Strategy Lockout\\
* 02:00 PM - 05:00 PM: Core Technical Debt Prioritization Workshop\\

Wednesday\\
* 09:00 AM - 05:00 PM: All-Day Emergency Security Operations Simulation (Do Not Interrupt)\\

Thursday\\
* 09:00 AM - 01:00 PM: Vendor Contract Renegotiation \& Legal Alignment\\
* 01:00 PM - 01:30 PM: [No events scheduled]\\
* 01:30 PM - 04:00 PM: Departmental Reorg \& Resource Allocation Review\\
* 04:00 PM - 05:00 PM: Immediate Escalation Meeting - Client Onboarding Block\\

Friday\\
* 09:00 AM - 12:00 PM: Global Compliance \& Privacy Audit Sign-off\\
* 12:00 PM - 12:30 PM: [No events scheduled]\\
* 12:30 PM - 04:30 PM: High-Stakes M\&A Due Diligence Deep Dive\\
* 04:30 PM - 05:00 PM: Q3 Product Roadmap Finalization Working Lunch\\
\end{tcolorbox}
\caption{}
\label{tab: consultant e schedule}
\end{table}

\section{Statistical Analysis, Plan Efficiency and Information Leakage}\label{app: statistical validation}

\subsection{Statistical Analysis}

\subsubsection{Experiment 1}
We first note that if GRASP outperforms its closest competitor in a statistically significant way, then it also statistically outperforms all other competitors. Therefore, we only look at the closest performers in this section.
\paragraph{ZebraLogic} We compare GRASP with PlanGEN(BoN) in this case. We use the McNemar test to determine that the p-value is $<0.0001$. Therefore, GRASP outperforms every baseline with high confidence from the statistical perspective.

\paragraph{Natural Plan Calendar Scheduling} We compare against PlanGEN(BoN) in this case as well. Even though GRASP outperforms PlanGEN(BoN) by $2\%$, by applying the McNemar test, we find that the result is not statistically significant with a p-value of $0.33$. However, GRASP outperforms all other baselines in a statistically significant fashion, with a p-value $<0.01$.

\paragraph{SciBench} Since each SciBench Math subset is a very small dataset, achieving statistical significance requires a significant increase in accuracy. However, due to the difficulty of the dataset, even a relatively modest increase is highly valuable. Therefore, we provide the bootstrapped $95\%$ confidence intervals for this dataset. For the SciBench Stat dataset, the confidence intervals are $[75\%,91.67\%]$ for GRASP and $[68.06\%, 87.5\%]$ for the other two methods. The p-value is $0.3438$ in this case. For SciBench Calc and Diff, the PlanGEN baselines perform at the same level as GRASP, so we do not perform a statistical analysis for them.

\paragraph{GPQA} We compare GRASP against the direct planner in this case, and find that the p-value is $0.6793$, which clarifies that GRASP does not outperform the direct gpt-4o planner in a statistically significant way.

\subsubsection{Experiment 2}
First, we observe that GRASP(GPT-4o) outperforms all other baselines with a p-value of $<0.01$ under the McNemar test. For the ablation study in Table~\ref{tab:7_way_ablation}, GRASP also outperforms all other models with a p-value of $<0.01$ when averaged over three runs.

\subsection{Plan Efficiency for Experiment 1}
\paragraph{ZebraLogic} The average plan length is 12.67 steps for GRASP and 9.74 steps for PlanGEN(BoN). This shows that the plans generated by GRASP are much longer. This tells us that GRASP autonomously generates longer plans for challenging problems, and the accuracy increases by a large margin as a result.

\paragraph{Natural Plan Calendar Scheduling} We find that the plans for GRASP average 11.3 steps, whereas for PlanGEN(BoN) they average 13.09 steps.

\paragraph{SciBench} For the SciBench Stat dataset, we find that the plans generated by GRASP average 9.18 steps, the plans for BoN average 6.96 steps and the plans for PlanGEN(Rebase) average 9.53 steps. For the SciBench Diff dataset, we find that the plans generated by GRASP average 11.66 steps and the plans generated by PlanGEN(Mix) average 14.2 steps. For the SciBench Calc dataset, we find that the plans generated by GRASP average 11.39 steps, and the plans generated by PlanGEN(Mix) average 14.85 steps.

\subsection{Information Leakage in Experiment 1}
We say that there is information leakage in a plan if the final answer for the task-instance pair appears anywhere in the plan.

\paragraph{ZebraLogic} We find that 31 plans with information leakage were generated by GRASP, whereas PlanGEN(BoN) generated 145 such plans. Therefore, if we calculate the accuracy while considering plans that violate the constraint that the final answer of the task instance pair should not be given as an output, then the accuracy of GRASP(gpt-4o-mini) remains $55.7\%$, whereas the accuracy of PlanGEN(BoN) falls to $27.1\%$. This shows that enforcing constraints significantly reduces the accuracy in the baseline methods, but GRASP shows very little plan information leakage, and follows constraints in a strict fashion.

\paragraph{Natural Plan Calendar Scheduling} We find that no plan generated by GRASP had information leakage, whereas $103$ plans generated by PlanGEN(BoN) had information leakage. Therefore, the accuracy of GRASP remains $74.3\%$, whereas the accuracy of PlanGEN(BoN) drops to $62\%$. This shows that when constraints are strictly enforced, GRASP outperforms all other methods by a significant margin. 

\paragraph{SciBench} For the SciBench Stat dataset, we find that there is one plan where information leakage happens in both GRASP and PlanGEN(Rebase), but no leakage happens in BoN. This leads to a revised accuracy of $80.56\%$ for GRASP, $77.78\%$ for BoN and $76.39\%$ for PlanGEN(Rebase). For the SciBench Diff dataset, we find that both GRASP and PlanGEN(Mix) have four plans where leakage happens. This leads to a revised accuracy of $42\%$ for both GRASP and PlanGEN(Mix). For the SciBench Calc dataset, we find that there are two plans where information leakage happens in GRASP, and seven plans where information leakage happens in PlanGEN(Mix). This leads to a revised accuracy of $68.27\%$ for GRASP, and $56.07\%$ for PlanGEN(Mix).

\subsection{Discussion}
When we look at the statistical analysis of the results from Section~\ref{sec: main results}, we see that some results are not statistically significant. However, upon closer inspection, we see that GRASP consistently generates shorter plans which yield higher accuracy, while consuming less tokens than state-of-the-art baselines. Moreover, when a simple constraint is strictly enforced, we see that GRASP maintains a similar accuracy, whereas the accuracies of baselines drop by more than $10\%$. This shows that GRASP achieves higher accuracy, generates shorter plans (which leads to downstream savings in token consumption as well), consumes less tokens, while maintaining constraints and accuracy under multi-task instructions.

\section{GenPlan Algorithm and Prompts}\label{app:genplan}
The algorithm for GenPlan is given in Algorithm~\ref{alg:GenPlan}. The individual prompts are given below.

\begin{algorithm}[h]
\caption{GenPlan}
\label{alg:GenPlan}
\footnotesize
\begin{algorithmic}
\REQUIRE Task $\mathcal{T}$, User Specified Guidelines $\mathcal{G}_{\text{user}}$, User Specified Constraints $\mathcal{C}_{\text{user}}$, $\text{MAX\_ITER}$
    \ENSURE Finalized Baseline Plan $\mathcal{P}_G$
    
    \IF{$\mathcal{G}_{\text{user}} = \emptyset$}
        \STATE $\mathcal{G}_0 \gets \text{GA}(\mathcal{T})$
    \ELSE
        \STATE $\mathcal{G}_0 \gets \mathcal{G}_{\text{user}}$
    \ENDIF
    
    \IF{$\mathcal{C}_{\text{user}} = \emptyset$}
        \STATE $\mathcal{C}_0 \gets \text{CA}(\mathcal{T})$
    \ELSE
        \STATE $\mathcal{C}_0 \gets \mathcal{C}_{\text{user}}$
    \ENDIF
    
    \STATE $\mathcal{P}_0 \gets \emptyset$
    \STATE $\mathcal{KB} \gets \langle \mathcal{T}, \mathcal{C}_0, \mathcal{G}_0, \mathcal{P}_0 \rangle$
    
    \STATE $guidelines\_flag \gets \text{TRUE}$
    \STATE $constraints\_flag \gets \text{TRUE}$
    \STATE $j \gets 0$
    
    \WHILE{$(guidelines\_flag = \text{TRUE} \textbf{ or } constraints\_flag = \text{TRUE}) \textbf{ and } j \le \text{MAX\_ITER}$}
        \STATE $j \gets j + 1$
        
        \STATE $\mathcal{G}_{\text{prev}} \gets \mathcal{KB}[\text{`Guidelines'}]$
        \STATE $\mathcal{C}_{\text{prev}} \gets \mathcal{KB}[\text{`Constraints'}]$
        
        \STATE $\mathcal{G}_{\text{new}} \gets \text{GA}(\mathcal{KB})$
        \STATE $\mathcal{C}_{\text{new}} \gets \text{CA}(\mathcal{KB})$
        
        \STATE $\mathcal{G}_{\text{merged}} \gets \text{MERGE}(\mathcal{G}_{\text{prev}}, \mathcal{G}_{\text{new}})$
        \STATE $\mathcal{C}_{\text{merged}} \gets \text{MERGE}(\mathcal{C}_{\text{prev}}, \mathcal{C}_{\text{new}})$

        \STATE $\mathcal{KB}[\text{`Plan'}] \gets \text{PGA}(\mathcal{KB})$
        
        \STATE $\mathcal{KB}[\text{`Guidelines'}] \gets \mathcal{G}_{\text{merged}}$
        \STATE $\mathcal{KB}[\text{`Constraints'}] \gets \mathcal{C}_{\text{merged}}$
        
        \IF{$\text{IsEquivalent}(\mathcal{G}_{\text{prev}}, \mathcal{G}_{\text{merged}})$}
            \STATE $guidelines\_flag \gets \text{FALSE}$
        \ENDIF
        
        \IF{$\text{IsEquivalent}(\mathcal{C}_{\text{prev}}, \mathcal{C}_{\text{merged}})$}
            \STATE $constraints\_flag \gets \text{FALSE}$
        \ENDIF
    \ENDWHILE
    
    \STATE $\mathcal{P}_G \gets \mathcal{KB}[\text{'Plan'}]$
    \RETURN $\mathcal{P}_G$
\end{algorithmic}
\end{algorithm}

\begin{tcolorbox}[colback=gray!10, colframe=black, title=Merge\_Agent Prompt]
\footnotesize messages=[\{"role":"system","content":"""You are an expert in organizing and consolidating information.
                        You will be given an existing list and a new list. Your task is to merge them into a single coherent numbered list.\\
                        Rules:\\
                        1. Remove any duplicate or redundant points\\
                        2. Combine similar points into single, comprehensive ones\\
                        3. Maintain logical flow and organization\\
                        4. Output only the numbered list without any headers or additional text\\
                        5. Ensure all unique and important points from both inputs are included\\
                        6. Use clear, concise language for each guideline\\
        7. If the new list is empty, return the existing list."""\},\\
                        \{"role":"user","content":f"Existing list:\textbackslash n{existing\_list}\textbackslash n\textbackslash nNew list:\textbackslash n{new\_list}\textbackslash n\textbackslash nPlease merge these into a single coherent numbered list:"\}]
\end{tcolorbox}

\begin{tcolorbox}[colback=gray!10, colframe=black, title=is\_equivalent Prompt]
\footnotesize messages=[
                \{"role": "system", "content": """You are an expert at comparing two numbered lists for semantic equivalence. You will be given two numbered lists as plain text. Your job is to determine if they contain essentially the same information, even if the wording or order is different. If the lists are equivalent in meaning, respond with only 'Yes'. If they are not, respond with only 'No'. Do not provide any explanation or extra text."""\},
                \{"role": "user", "content": f"""List 1:\textbackslash n{list1\_str}\textbackslash n\textbackslash nList 2:\textbackslash n{list2\_str}\textbackslash n\textbackslash nAre these two lists essentially the same in terms of the information they contain?"""\}
            ]
\end{tcolorbox}

\begin{tcolorbox}[colback=gray!10, colframe=black, title=Guidelines\_Agent Prompt]
\footnotesize messages = [
                            \{
                                "role": "system",
                                "content": """You are an expert in designing effective guidelines for solving tasks in a general and reusable way.\\
                                You will be provided with a Knowledge Base that may include:\\
                                - Existing guidelines\\
                                - Constraints\\
                                - A task description\\
                                - An instance of the task\\
                                Your task is to:\\
                                1. Analyze the entire knowledge base carefully.\\
                                2. Identify and generate **new general-purpose guidelines** that are relevant to solving the task, not just for a specific input.\\
                                3. These guidelines may include best practices, strategic approaches, or helpful techniques.\\
                                5. If there are no new guidelines, output the existing ones unchanged.\\
                                6. Make sure that the guidelines do not suggest ideas that are not related to solving the task and instance.\\

                                Output format:\\
                                - A **numbered list** of guidelines, as plain text (no headers).\\

                                Ensure that the following guidelines are included in the list of guidelines:\\
                                1. Each step of the plan must clearly describe what needs to be done, providing all necessary information required to implement the step.\\
                                2. The output of a step should not be given in the step. For example, the step cannot be "Find the answer to the question: answer".\\
                                3. If a step depends on the result of a previous step, refer to that step by its step number (e.g., "use the result from Step 3") — do not reference or include the output value itself.\\
                                4. It correctly provides directions to solve the task for the given instance.\\
                                5. It should not provide any solutions in each step. It should only provide the complete directions to implement the step.\\
                                6. The final solution should not appear in the plan.\\
                                7. Each step should include all the information necessary to implement the step, except the final output of the step.\\
                                8. The steps should not bias the LLM to a particular solution.\\

                                If the knowledge base contains no existing guidelines, you may return "No guidelines added.".
                                """
                            \},\\
                            \{
                                "role": "user",
                                "content": f"Knowledge Base: \{KNOWLEDGE\_BASE\}"
                            \}
                            ]
\end{tcolorbox}

\begin{tcolorbox}[colback=gray!10, colframe=black, title=Constraints\_Agent Prompt]
\footnotesize messages = [
                \{
                    "role": "system",
                    "content": """You are an expert in understanding problems and formulating logical constraints.

                    You will be provided with a Knowledge Base that may include:\\
                    - Existing constraints\\
                    - Guidelines\\
                    - Facts\\
                    - A task description\\
                    - An instance of the task\\

                    Your task is to:\\
                    1. Carefully analyze the entire knowledge base.\\
                    2. Extract and generate any new constraints that are logically required, including:\\
                    - Constraints that are implied but not explicitly stated\\
                    - Assumptions necessary for interpreting or solving the problem\\
                    3. Try not to repeat constraints already in the knowledge base.\\
                    4. If no new constraints are needed, return "No constraints added."\\.

                    Output format:
                    - A **numbered list** of constraints, as plain text (no headers).
                    """
                \},\\
                \{
                    "role": "user",
                    "content": f"Knowledge Base: \{KNOWLEDGE\_BASE\}"
                \}
                ]
\end{tcolorbox}

\begin{tcolorbox}[colback=gray!10, colframe=black, title=Plan\_Generation\_Agent Prompt]
\footnotesize messages=[\\\{"role":"system","content":"""You are an expert in understanding a problem and generating a plan to solve the problem.
            You are given a task that needs to be carried out in the knowledge base. 
            You will modify the current plan to solve the problem.
            The knowledge base given by the user already contains a list of guidelines and constraints.
            Analyze the input knowledge base and modify the current plan to solve the problem.
            Each step in the plan should be a single sentence without any headers.
            You will output the plan as a numbered list without any other text."""\},\\\{"role":"user","content":f"Knowledge Base: \{KNOWLEDGE\_BASE\}"\}]
\end{tcolorbox}

\section{RevPlan Algorithm and Prompts}\label{app:revplan}

The algorithm for RevPlan is given in Algorithm~\ref{alg:RevPlan}. The individual prompts are given below.

\begin{algorithm}
\caption{RevPlan}
\label{alg:RevPlan}
\footnotesize
\begin{algorithmic}
    \REQUIRE Task Description $\mathcal{T}$, Task Instance $\mathcal{I}$, Baseline Plan $\mathcal{P}_G$\\
    \ENSURE Set of Specialized Candidate Plans $\mathbb{P}$\\
    
    \STATE $\mathbb{P} \gets \emptyset$\\
    \STATE $\mathcal{S} \gets \text{GenerateStrategies}(\mathcal{T}, \mathcal{I})$\\
    
    \FOR{each strategy $s_k \in \mathcal{S}$}
        \STATE $\mathcal{C}_{s_k} \gets \text{GenerateLocalConstraints}(s_k, \mathcal{I})$\\
        
        \STATE $\chi_{s_k} \gets \text{RunReAct}(\mathcal{T}, \mathcal{I}, s_k, \mathcal{C}_{s_k})$\\
        
        \STATE $\mathcal{P}_{\text{specific}}^{(k)} \gets \text{ExtractPlan}(\chi_{s_k})$\\
        
        \STATE $\mathcal{P}^{(k)} \gets \text{MergePlans}(\mathcal{P}_{\text{specific}}^{(k)}, \mathcal{P}_G)$\\
        
        \STATE $\mathbb{P} \gets \mathbb{P} \cup \{\mathcal{P}^{(k)}\}$\\
    \ENDFOR
    \STATE \RETURN $\mathbb{P}$
\end{algorithmic}
\end{algorithm}

\begin{tcolorbox}[colback=gray!10, colframe=black, title=merge\_plans Prompt]
\footnotesize You are an expert in fitting a plan to a given plan template.\\
                You are given a plan template: \\\{KNOWLEDGE\_BASE('PLAN')\}\\
                You are given a plan: \{specific\_plan\}\\
                You need to fit the plan to the plan template loosely, which means you can add, remove, or modify the plan template to fit the plan in such a way that the plan is more robust.\\
                Please output the plan fitted to the plan template as a numbered list of steps. without any other text.
\end{tcolorbox}

\begin{tcolorbox}[colback=gray!10, colframe=black, title=Strategy-specific constraints generation Prompt]
\footnotesize You are an expert at generating boundary constraints and implicit assumptions for a plan.\\
            I will give you:\\

            - A problem description.\\
            - A high-level strategy to solve it.\\
            - An overarching task description that is the overarching task associated with the problem.\\

            Your task is to list all explicit constraints that the final plan must satisfy in order to be valid and effective.\\

            Overarching task: \{task\}\\
            Instance: \{instance\}\\
            Strategy: \{strategy\}\\
            Strategy Details: \{strategy\_principle\} \{strategy\_explanation\}\\

            Output: A numbered list of constraints and implicit assumptions. Do not include any other text.
\end{tcolorbox}

\begin{tcolorbox}[colback=gray!10, colframe=black, title=extract\_plan Prompt]
\footnotesize You are an expert at extracting core steps from a solution to a problem while ignoring minute implementation details and ignoring the results of the steps.\\
                You will be given a ReAct based thought process to a problem and a strategy. You need to extract the core steps from the thought process while ignoring minute implementation details and results of the steps.
                If the information is not present directly in the problem or task description, then do not include it in the plan.\\

                The overarching task associated with the problem is: \{task\}\\
                The problem is: \{instance\}\\

                The ReAct based thought process and solution with details is: \{ReAct\_solution\}\\

                The strategy is:\\
                \{strategy\}, Strategy Details: \{strategy\_principle\} \{strategy\_explanation\}\\
                Please output the plan as a numbered list of steps without any other text.\\

                NOTE: Do not include the results of the steps in the plan.
\end{tcolorbox}

\begin{tcolorbox}[colback=gray!10, colframe=black, title=Generate\_Strategies Prompt]
\footnotesize messages=[\{"role": "system", "content": "You are an expert in coming up with strategies to solve problems. The strategies you come up with are solvable by an LLM and do not require any external tools or resources."\},\\\{"role": "user", "content": f"""Here is the problem: \{instance\}\\

Please provide exactly \{NUM\_STRATEGIES\} different primary THEORETICAL approaches to solve this problem. Each approach should be distinct and based on a different principle, method, or analytical perspective, covering a range of techniques if possible.

Each of the \{NUM\_STRATEGIES\} approaches should be implementable by an LLM with strong logical reasoning capabilities.

For each of the \{NUM\_STRATEGIES\} approaches:\\
- Clearly name or title the approach.\\
- Highlight the key principle behind it and provide a brief explanation of how it works and why it could be effective for this type of problem.\\
- Provide the strategy involved in solving the problem.\\

For each strategy, follow the following format:\\
- Principle: <principle>\\
- Explanation: <explanation>\\

The overarching task associated with the problem is: \{task\}\\

Present your answer as a numbered list of length \{NUM\_STRATEGIES\} for clarity. Do not include any other text and follow the format exactly.
"""\}]
\end{tcolorbox}

\section{VerPlan Algorithm and Prompts}

The algorithm for VerPlan is given in Algorithm~\ref{alg:VerPlan}. The prompt for VerPlan is given below.

\begin{algorithm}[h]
\caption{VerPlan}
\label{alg:VerPlan}
\footnotesize 
\begin{algorithmic}

    \REQUIRE Task Description $\mathcal{T}$, Task Instance $\mathcal{I}$, Constraints $\mathcal{C}$, Set of Candidate Plans $\mathbb{P} = \{\mathcal{P}^{(1)}, \mathcal{P}^{(2)}, \ldots, \mathcal{P}^{(K)}\}$, Set of Strategy-specific Constraints, $\{\mathcal{C}_{s_1}, \ldots, \mathcal{C}_{s_K}\}$, Rubrics
    \ENSURE Final Plan $\mathcal{P}^*$
    
    \STATE $\Psi_{\text{best}} \gets -1$
    \STATE $\mathcal{P}^* \gets \emptyset$
    
    \FOR{each candidate plan $\mathcal{P}^{(k)} \in \mathbb{P}$}
        \STATE $\Psi^{(k)} \gets \text{Score}\left(\mathcal{P}^{(k)} \mid \mathcal{T}, \mathcal{I}, \mathcal{C}, \mathcal{C}_{s_k}, \text{Rubrics}\right)$
        
        \IF{$\Psi^{(k)} > \Psi_{\text{best}}$}
            \STATE $\Psi_{\text{best}} \gets \Psi^{(k)}$
            \STATE $\mathcal{P}^* \gets \mathcal{P}^{(k)}$
        \ENDIF
    \ENDFOR
    
    \RETURN $\mathcal{P}^*$

\end{algorithmic}
\end{algorithm}

\begin{tcolorbox}[colback=gray!10, colframe=black, title=VerPlan Prompt]
\footnotesize Rate this plan on a scale of 1-100 based on how well it solves the given problem.

Problem: \{problem\_text\}

Constraints: \{strategy\_specific\_constraints\}

Plan: \{strategy\_specific\_plan\}

Overarching task: \{task\}

Consider:
- How complete is the plan?
- How likely is it to solve the problem correctly?
- How clear and actionable are the steps?
- How well does it address the requirements of the problem?

Respond with only a single integer from 1-100.
\end{tcolorbox}

\section{Direct Planner Generation Prompt}\label{app: gpt-4o plan generation prompt}

\begin{tcolorbox}[colback=gray!10, colframe=black, title=Direct Planner Generation Prompt: Experiment 1]
\footnotesize You are an expert planner.
                    You will be given a TASK and an INSTANCE of the task.
                    Your job is to produce an executable plan as a numbered list.
                    
                    Rules:
                    - Output ONLY the numbered list. No other text.\newline
                    - Do NOT include any intermediate results, computed values, or the final answer inside any step.\newline
                    - Steps should be actionable and reference prior steps by step number when needed.\newline
                    - The plan should give the final answer as the output.\newline
                    <task> \{task\}
</task>

<instance>
\{instance\}
</instance>

Generate an instance-specific plan to solve the task for this instance.
Output only the plan as a numbered list.
\end{tcolorbox}

\begin{tcolorbox}[colback=gray!10, colframe=black, title=Direct Planner Generation Prompt: Experiment 2]
\footnotesize You are an expert planner.\\
You will be given a TASK and an INSTANCE of the task.\\
Your job is to produce an executable plan as a numbered list.\\

Rules:\\
- A feasible solution always exists.\\
- Output ONLY the numbered list. No other text.\\
- Do NOT include any intermediate results, computed values, or the final answer inside any step.\\
- Steps should be actionable and reference prior steps by step number when needed.\\
- The final step of the plan should be an instruction to give the final answer as the output. The final solution should not appear in the plan.\\
<task> \{task\}</task>\\

<instance>
\{instance\}
</instance>\\

Generate a plan to solve the task for this instance. Ensure that the steps can be used to find the final answer organically, rather than using your own calculation.\\
Output only the plan as a numbered list.
\end{tcolorbox}

\section{Examples of Guidelines and Constraints}\label{app: guidelines and constraints}
We share an example of guidelines and constraints for Natural Plan Calendar Scheduling in Table~\ref{tab:calendar-guidelines} and Table~\ref{tab:calendar-constraints}. We share an example of guidelines and constraints for ZebraLogic in Table~\ref{tab:zebralogic-guidelines} and Table~\ref{tab:Zebralogic-constraints}.

\begin{table}[htbp]
\centering
\begin{tcolorbox}[colback=gray!10, colframe=black, title= Guidelines (Natural Plan Calendar Scheduling example: Dual Task)]
\footnotesize 1. Each step of the plan must clearly describe what needs to be done, providing all necessary information required to implement the step.\\
2. The output of a step should not be given in the step. For example, the step cannot be "Find the answer to the question: answer".\\  
3. If a step depends on the result of a previous step, refer to that step by its step number (e.g., "use the result from Step 3") — do not reference or include the output value itself.\\
4. If you mention a guideline or constraint, explicitly word it out in the step verbatim.\\
5. Identify the primary group meeting schedule requirements, including the day and time range, and ensure the format matches "Day, HH:MM AM/PM - HH:MM AM/PM".\\
6. Gather all necessary information about the primary group meeting, such as availability, preferences, and constraints, and determine the final schedule.\\
7. Analyze the 5-day schedule of External Consultant D, ensuring the working hours are limited to Monday through Friday, 9:00 AM to 5:00 PM.\\
8. Identify all free intervals in External Consultant D's schedule that are at least 30 minutes long, and determine the absolute latest interval.\\
9. Output only the day and start time of the latest 30-minute free interval in the format "Day, HH:MM AM/PM".\\
10. Ensure the final response consists of exactly two lines, with the first line corresponding to Task 1 and the second line corresponding to Task 2, separated by a newline character.\\
11. Verify that the output adheres strictly to the required formats and constraints before finalizing the response.
\end{tcolorbox}
\caption{}
\label{tab:calendar-guidelines}
\end{table}

\begin{table}[htbp]
\centering
\begin{tcolorbox}[colback=gray!10, colframe=black, title= Constraints (Natural Plan Calendar Scheduling example: Dual Task)]
\footnotesize 1. The final response must consist of exactly two lines, separated by a newline character, with no additional conversational text, intermediate steps, or explanations.\\
2. The final answer should be given as the output without any additional text or explanation.\\
3. The format for the primary group meeting schedule (Task 1) must strictly follow: "Day, HH:MM AM/PM - HH:MM AM/PM".\\
4. The format for the latest 30-minute free interval for External Consultant D (Task 2) must strictly follow: "Day, HH:MM AM/PM".\\
5. The working hours for External Consultant D are Monday through Friday, 9:00 AM to 5:00 PM.\\
6. The latest 30-minute free interval for Task 2 must be determined within the standard working week and working hours of External Consultant D.\\
7. The analysis for Task 2 must consider the absolute latest free interval, meaning the interval closest to the end of the working hours on Friday, if available.
\end{tcolorbox}
\caption{}
\label{tab:calendar-constraints}
\end{table}

\begin{table}[htbp]
\centering
\begin{tcolorbox}[colback=gray!10, colframe=black, title= Guidelines (ZebraLogic example)]
\footnotesize 1. Carefully read all given clues and constraints to fully understand the requirements and limitations of the task.\\
2. Clearly describe each step in the plan, providing all necessary information required to implement it without mentioning the output within the step.\\
3. Identify key elements and variables pertinent to the task, capturing all details and constraints accurately.\\
4. List clues sequentially, ensuring accuracy and completeness in capturing constraints and requirements.\\
5. Develop a logical deduction plan, creating step-by-step instructions to address each part of the task while considering all constraints.\\
6. Validate consistency and accuracy by cross-referencing each deduction step with the clues.\\
7. If a step depends on prior results, refer to the output from previous steps by step number without including the output value itself.\\
8. Once deductions are verified, identify the optimal solution that aligns with all clues and constraints.\\
9. Format the final solution according to the task specifications, ensuring clarity and correctness.\\
10. For complex logical deductions, implement them in Python with appropriate use of loops, conditionals, and tracking, ensuring precision.\\
11. Re-check each step and the final solution for adherence to the specified format and constraints.\\
12. Explicitly word out any guidelines or constraints in the steps as needed.
\end{tcolorbox}
\caption{}
\label{tab:zebralogic-guidelines}
\end{table}

\begin{table}[htbp]
\centering
\begin{tcolorbox}[colback=gray!10, colframe=black, title= Constraints (ZebraLogic example)]
\footnotesize 1. The logical riddle may involve a sequence of steps that require interpreting clues accurately, whether provided implicitly or explicitly.\\
2. Ensure each step logically flows from the previous one, using the outcome of each prior step as the input for the next, without contradicting specified constraints.\\
3. All assumptions must be consistent with the guidelines and existing constraints; derived constraints should not alter the given question or its fundamental goal.\\
4. Only logical reasoning should be utilized for solving the task, without changing the question.\\
5. Ensure that the final answer is the only deliverable, presented in the specified output format without any introductory or explanatory text.
\end{tcolorbox}
\caption{}
\label{tab:Zebralogic-constraints}
\end{table}

\section{Plan Revision}\label{app:II}

In this section, we present two exemplary scenarios on the plan revision that is made by the RevPlan agent. First, we consider Calendar Scheduling from the Natural Plan dataset \cite{zheng2024natural}. The task description and instance of the task are given in Table~\ref{tab:calendar-task}. For this task and a given instance of scheduling a meeting with constraints, the initial generated plan is outlined in Table~\ref{tab:calendar-plan}. The RevPlan agent first generates possible strategies to construct feasible schedules. The structure of the suggested strategies comprises of at least a \emph{principle} and \emph{explanation}. The two suggested strategies and the revised plan based on each strategy are given in Table~\ref{tab:calendar-plan1} and Table~\ref{tab:calendar-plan2}. Next, we provide another example of plan revision by the RevPlan agent. The task and the specific instance of the task are taken from the GPQA dataset and are given in Table~\ref{tab:gpqa-task}. The initial plan generated by the GenPlan agent is shown in Table~\ref{tab:gpqa-plan}. By setting the number of strategies to four, for this task and instance, we automatically generate 4 different strategies in GRASP, and the RevPlan agent revises the initial plan according to each plan. Table~\ref{tab:gpqa-all-strategies} present the strategies and their corresponding refined plans.

\begin{table}[htbp]
\centering
\begin{tcolorbox}[colback=gray!10, colframe=black, title= Task and Instance Description (Natural Plan Calendar Scheduling example)]
\footnotesize

\textbf{Task:} \\
Your task is to find a feasible schedule for calendar events, taking into account time slots, durations, constraints, and conflicts. You will analyze the calendar requirements and available time slots to determine the optimal event schedule. Output the final calendar schedule with all relevant details about the events and times, without any other text.
\vspace{0.5em}

\textbf{Instance:} \\
You need to schedule a meeting for Thomas, Dylan and Jerry for one hour between the work hours of 9:00 to 17:00 on Monday. $\backslash$ n $\backslash$ nHere are the existing schedules for everyone during the day: $\backslash$ nThomas's calendar is wide open the entire day.$\backslash$ nDylan has blocked their calendar on Monday during 10:30 to 11:00, 13:30 to 14:00; $\backslash$ nJerry is busy on Monday during 9:00 to 11:00, 11:30 to 14:30, 16:00 to 17:00; $\backslash$ n$\backslash$ nFind a time that works for everyone's schedule and constraints.
\end{tcolorbox}
\caption{}
\label{tab:calendar-task}
\end{table}

\begin{table}[h!]
\centering
\begin{tcolorbox}[colback=gray!10, colframe=black, title= Initial plan generated by GenPlan agent]
\footnotesize

\begin{enumerate}\itemsep0em
\item Analyze the calendar requirements and available time slots to list all potential events and their durations.
\item Identify any conflicts and constraints among the events and time slots using the information from the constraints section.
\item Calculate the number of available days for each event by examining the time slots and durations, explicitly counting the number of days spent as per constraints.
\item Determine the sequence of events by considering the dependencies and constraints, ensuring the last day of departure is the same as the first day of arrival at the next event.
\item Allocate the calculated days to each event sequence, ensuring the total days and flow from event to event follow the constraints.
\item Verify the allocation meets all specified constraints by checking conditions about the day count and departure/arrival continuity.
\item Compile the final schedule by arranging the events with their respective days and durations while ensuring they adhere to all constraints.
\item Output the final calendar schedule with all relevant details about the events and times, without any other text.
\end{enumerate}
\end{tcolorbox}
\caption{Initial plan generated by GenPlan agent for Natural Plan (Calendar Scheduling)}
\label{tab:calendar-plan}
\end{table}

\begin{table}[htbp]
\centering
\begin{tcolorbox}[colback=gray!10, colframe=black, title= Refined Plan revised by RevPlan agent]
\footnotesize

\textbf{1. Interval Analysis} \\
\textit{Principle:} Interval Intersection \\
\textit{Explanation:} This approach involves representing each person's schedule as a series of time intervals (busy times) and looks to find a one-hour window (free time) that intersects across all individuals involved. By identifying the gaps between the busy intervals for each participant, the task then evaluates these gaps to find a common free interval. This method is effective as it directly focuses on usable time slots rather than entire schedules.
\\

\textbf{Revised Plan:}
\begin{enumerate}
\item Analyze the calendar requirements and identify the busy intervals for Dylan and Jerry.
\item Identify constraints and dependencies, ensuring the intended meeting includes Thomas's availability.
\item Determine Dylan's and Jerry's free intervals by examining their respective busy schedules.
\item Compare Dylan's and Jerry's free intervals to find all common available one-hour slots.
\item Check these common slots against the constraint of matching Thomas's availability for final time slot selection.
\item Ensure last-minute adjustments consider any changes in individual schedules while maintaining the main constraints.
\item Compile a draft schedule highlighting all potential common slots, specifying the chosen slot that suits all three individuals.
\item Output the final schedule for the proposed one-hour meeting, listing the confirmed time and date.
\end{enumerate}

\end{tcolorbox}
\caption{Revised Plan Based on Strategy 1 (Calendar Scheduling example, instance \#6)}
\label{tab:calendar-plan1}
\end{table}

\begin{table}[ht]
\centering
\begin{tcolorbox}[colback=gray!10, colframe=black, title= Refined Plan revised by RevPlan agent]
\footnotesize
\textbf{2. Slot Evaluation} \\
\textit{Principle:} Discrete Slot-Checking \\
\textit{Explanation:} This method discretizes the day into a series of potential one-hour meeting slots and iteratively checks each slot against the calendars of all participants. By scanning through these slots one by one and verifying availability, it finds the first viable time. This approach is effective in systematically ensuring that each potential time is considered, minimizing chances of oversight.
\\

\textbf{Revised Plan:}
\begin{enumerate}
\item Analyze the calendar for Monday from 9:00 to 17:00 and list all potential one-hour time slots.
\item Identify any conflicts by checking Dylan's and Jerry's unavailability against these time slots.
\item Calculate the number of available one-hour slots where all participants are free.
\item Determine the sequence of available slots, ensuring a continuous flow without conflicts.
\item Select the first available time slot where all participants (Dylan and Jerry) are available.
\item Verify the selected time slot meets all specified constraints of availability for both participants.
\item Compile the final schedule by arranging the selected time slot and confirming participant availability.
\item Output the final time slot schedule with all relevant details about participant availability.
\end{enumerate}

\end{tcolorbox}
\caption{Revised Plan Based on Strategy 2 (Calendar Scheduling example, instance \#6)}
\label{tab:calendar-plan2}
\end{table}

\begin{table}[htbp]
\centering
\begin{tcolorbox}[colback=gray!10, colframe=black, title= Task and Instance Description (GPQA example)]
\footnotesize

\textbf{Task:} \\
You are an expert at reasoning and solving graduate-level science multiple choice questions across physics, chemistry, and biology domains. Please answer the question given in the input text by providing detailed reasoning and the final answer, which is the letter choice (A, B, C, D) of the correct answer.
\vspace{0.5em}

\textbf{Instance:} \\
Two quantum states with energies \(E_1\) and \(E_2\) have lifetimes of \(10^{-9}\) sec and \(10^{-8}\) sec, respectively. We want to clearly distinguish these two energy levels. Which one of the following options could be their energy difference so that they can be clearly resolved?
\end{tcolorbox}
\caption{Task and Instance for GPQA Plan Revision}
\label{tab:gpqa-task}
\end{table}

\begin{table}[htbp]
\centering
\begin{tcolorbox}[colback=gray!10, colframe=black, title= Initial plan generated by GenPlan agent]
\footnotesize

\begin{enumerate}
\item Read the entire question and all options A–D, record every given value with units, translate any figures/tables into explicit variables, constraints, or relationships, identify the scientific domain and subtopic, and state precisely what quantity or concept is being asked. 
\item Define symbols for all knowns and unknowns with SI units, choose and state coordinate systems, sign conventions, and reference states, list applicable principles/laws/mechanisms, declare justified assumptions and their domains of validity (e.g., ideal gas, dilute solution, nonrelativistic, small-angle), identify information gaps and minimal reasonable assumptions, and specify standard constants with selected numerical values and units to be used consistently (e.g., $g=9.81 m\cdot s^{-2}, R=8.314462618 J\cdot mol^{-1}\cdot K^{-1}, NA=6.02214076\times 10^{23} mol^{-1}, kB=1.380649\times 10^{-23} J\cdot K^{-1}, h=6.62607015\times 10^-34 J \cdot s, c=2.99792458\times 10^8 m\cdot s^{-1}, e=1.602176634\times 10^{-19} C, \varepsilon_0=8.854187817\times 10^{-12} F \cdot m^{-1}, \mu_0=4\pi\times 10^{-7} N \cdot A^{-2}, \sigma=5.670374419\times 10^{-8} W\cdot m^{-2} \cdot K^{-4}$), adding any field-specific reference values at T=298 K unless otherwise specified. 
\item Translate the problem into governing equations, logical inferences, or mechanistic pathways that connect the defined variables to the target quantity, enumerate the exact sequence of computations and unit conversions to be performed and which quantities will be computed, confirm dimensional homogeneity and unit consistency of every equation before use, and plan limiting-case or order-of-magnitude checks (and, for statistical/experimental design items, specify appropriate controls, confounders, test statistics, and assumptions). 
\item In the reasoning section, carry out the planned algebraic substitutions and arithmetic while tracking units and citing the relevant setup steps by number, and then compare the implications from the setup against each option A–D explicitly by checking signs, units, magnitudes, trends, stoichiometry, or mechanistic plausibility to eliminate incompatible choices. 
\item Perform an independent cross-check using a complementary principle that does not reuse the primary equations from Step 3 (e.g., energy vs. force balance, $\Delta G$ vs. K, oxidation states vs. stoichiometry, pathway logic vs. phenotype), reconcile any discrepancies, verify that the tentative option satisfies all assumptions and consistency checks from Steps 2–3 including unit and limiting-case validations, and then present the final answer on its own line as a single uppercase letter from {A, B, C, D}.
\end{enumerate}
\end{tcolorbox}
\caption{Initial plan generated by GenPlan agent for GPQA dataset}
\label{tab:gpqa-plan}
\end{table}

\begin{table*}[t]
\centering
\begin{tcolorbox}[
    colback=gray!8,
    colframe=black,
    title=Refined Planning Strategies for the GPQA Example,
    fonttitle=\bfseries,
    width=\textwidth,
    boxrule=0.8pt
]
\footnotesize

\textbf{1. Uncertainty Principle Approach} \\
\textit{Principle:} Energy-Time Uncertainty Principle \\
\textit{Explanation:} 
The relation $\Delta E \cdot \Delta t \geq \hbar/2$ implies that short-lived states exhibit larger energy uncertainty. To resolve two states, their energy separation must exceed the larger uncertainty.

\textbf{Refined Plan:}
\begin{enumerate}\itemsep0em
\item Identify the problem as an application of the uncertainty principle.
\item Apply $\Delta E \cdot \Delta t \geq \hbar/2$.
\item Compute $\Delta E$ for $10^{-9}$ s and $10^{-8}$ s.
\item Determine the larger uncertainty.
\item Compare with answer choices.
\item Select the smallest option exceeding the threshold.
\end{enumerate}

\vspace{0.8em}
\hrule
\vspace{0.8em}

\textbf{2. Spectroscopic Resolution Approach} \\
\textit{Principle:} Spectral Line Width and Resolution \\
\textit{Explanation:} 
Spectral line width is inversely proportional to lifetime. Two levels are resolvable only if their energy separation exceeds the broader line width.

\textbf{Refined Plan:}
\begin{enumerate}\itemsep0em
\item Recognize spectral resolution context.
\item Estimate $\Delta E$ using $\hbar/\Delta t$.
\item Compute uncertainties for both states.
\item Identify the dominant width.
\item Eliminate options below this threshold.
\item Choose the smallest valid remaining option.
\end{enumerate}

\vspace{0.8em}
\hrule
\vspace{0.8em}

\textbf{3. Frequency Difference Approach} \\
\textit{Principle:} Energy–Frequency Relation ($E = h\nu$) \\
\textit{Explanation:} 
Energy uncertainty corresponds to frequency width. To resolve two states, their frequency separation must exceed combined spectral widths.

\textbf{Refined Plan:}
\begin{enumerate}\itemsep0em
\item Convert lifetime to frequency width.
\item Translate frequency width to energy uncertainty.
\item Combine uncertainties if needed.
\item Compare with answer choices.
\item Select the smallest option above the minimum resolvable difference.
\end{enumerate}

\vspace{0.8em}
\hrule
\vspace{0.8em}

\textbf{4. Lifetime Ratio Analysis} \\
\textit{Principle:} Relative Lifetime Comparison \\
\textit{Explanation:} 
Shorter lifetimes produce broader uncertainties. Comparing the $10{:}1$ lifetime ratio helps determine the dominant energy spread.

\textbf{Refined Plan:}
\begin{enumerate}\itemsep0em
\item Compute $\Delta E \approx \hbar/\Delta t$ for both states.
\item Identify the dominant uncertainty.
\item Use the $10{:}1$ ratio for proportional reasoning.
\item Eliminate insufficient answer choices.
\item Choose the smallest option exceeding the larger uncertainty.
\end{enumerate}

\end{tcolorbox}
\caption{Refined Planning Strategies for the GPQA Example}
\label{tab:gpqa-all-strategies}
\end{table*}

\end{document}